\documentclass[authoryear]{elsarticle}

\makeatletter
\nopreprintlinetrue
\makeatother

\usepackage{bbm}
\usepackage[table]{xcolor}
\usepackage{graphicx}
\usepackage{soul, color}
\usepackage{rotating}
\usepackage{multirow}
\usepackage{booktabs}
\usepackage{amsfonts}
\usepackage{amsmath}
\usepackage{url}
\usepackage{colortbl}
\usepackage{lipsum}

\newcommand{\softplus}{\mathrm{softplus}}
\newcommand{\negBinp}{\text{p}}
\newcommand{\iid}{\overset{\mathrm{i.i.d.}}{\sim}}
\newcommand{\greytabline}{\rowcolor{backcolour}\cellcolor{white}}
\newcommand{\jmax}{j_{\mathrm{max}}}
\DeclareMathOperator*{\argmax}{arg\,max}
\newcommand{\ADIDA}{ADIDA$_{\mathrm{C}}$}
\newcommand{\RMSSE}{\textsc{rmsse}}
\newcommand{\CRPS}{\ensuremath{\textsc{rps}_{\text{0.5+}}}}

\definecolor{codeorange}{rgb}{0.9,0.3,0.}
\definecolor{codeblue}{rgb}{0.,0.5,0.99}
\definecolor{codepurple}{rgb}{0.58,0,0.82}
\definecolor{codegreen}{rgb}{0.,0.5,0.}
\definecolor{backcolour}{rgb}{0.95,0.95,0.92}
\definecolor{backcolour2}{rgb}{0.96,0.96,0.96}
\definecolor{codeblack}{rgb}{0.,0.,0.}

\newcommand{\Rev}[1]{\textcolor{codeblack}{#1}}
\newcommand{\Revnew}[1]{\textcolor{codeblack}{#1}}

\begin{document}

\begin{frontmatter}
\title{Forecasting intermittent time series with Gaussian Processes and Tweedie likelihood\tnoteref{special}\tnoteref{repr}}

\author{Stefano Damato\corref{cor1}}
\ead{stefano.damato@supsi.ch}

\author{Dario Azzimonti}
\ead{dario.azzimonti@supsi.it}

\author{Giorgio Corani}
\ead{giorgio.corani@supsi.ch}

\cortext[cor1]{Corresponding author}

\address{SUPSI, Istituto Dalle Molle di Studi sull'Intelligenza Artificiale (IDSIA),\\Lugano, Switzerland}

\tnotetext[special]{This article is part of a special issue entitled `ML/AI-Driven Forecasting in the Supply Chain' published in International Journal of Forecasting.}
\tnotetext[repr]{The numerical results presented in this manuscript were reproduced by CASCaD on 15 October 2025.}

\begin{abstract}
We \Rev{adopt} Gaussian Processes (GPs) \Rev{as latent functions} for probabilistic forecasting of intermittent time series. 
The model is trained in a Bayesian framework that accounts for the uncertainty about the latent function. 
We couple the latent GP variable with two types of forecast distributions: the negative binomial (NegBinGP) and the Tweedie distribution (TweedieGP). While the negative binomial has already been used in forecasting intermittent time series, this is the first time in which a fully parameterized Tweedie density is used for intermittent time series. We properly evaluate the Tweedie density, \Rev{which has both a point mass at zero and heavy tails}, avoiding simplifying assumptions made in existing models. 
We test our models on thousands of intermittent count time series. Results show that our models provide consistently better probabilistic forecasts than the competitors. 
In particular, TweedieGP obtains the best estimates of the highest quantiles, thus showing that it is more flexible than NegBinGP.
\end{abstract}

\end{frontmatter}

\newpage

\section{Introduction}
Intermittent time series characterize a large percentage of inventory items.
\Rev{Traditional} forecasting methods for intermittent demand \citep{Croston_1972, Syntetos_Boylan_2005,Nikolopoulos_Syntetos_Boylan_Petropoulos_Assimakopoulos_2011} provide point forecasts \Rev{only}. 
However, planning inventory levels
requires probability distributions from which to extract the relevant quantiles \citep{boylan2021intermittent,Kolassa_2016}.

Probabilistic models for intermittent \Rev{demand},
such as \citet[pp. 281-283]{Hyndman_2008}, \citet{snyder2012forecasting}, \citet{sbrana2023modelling}, and \citet{Svetunkov_Boylan_2023}, return the forecast distribution $p(y_{T+1} \mid y_{1:T})$, where $y_{1:T}$ are the observations available up to time $T$ and $y_{T+1}$ is the predicted value at time $T+1$.
Such models \Rev{predict} the distribution of the \textit{demand \Rev{size}} (the positive values of the time series) and the probability of \textit{occurrence}, i.e., the binary variable indicating whether there will be \Rev{positive} demand. 

\Revnew{Most probabilistic methods in the literature use one or more stochastic processes to control how the parameters of the forecast distribution change over time. Such stochastic processes are a function of time and their realisations are not observed, thus they are called latent.
For instance in the local level model \citep[Chap. 3.4]{Hyndman_2008},
the latent function parametrizes the mean of the forecast distribution.}

\Rev{Yet, the above models do not account for the uncertainty about the value of the
latent variables, which might be substantial. 
Indeed, \Revnew{incorporating} the uncertainty on} the parameters of the forecast distribution \Rev{leads to more reliable forecasts \citep{prak2019general}; this is a typical advantage of the Bayesian approach \citep[Chap.~9]{Gelman_Hill_Vehtari_2020}.} 

We propose to model the latent variable with a Gaussian Process (GP), a Bayesian non-parametric model which quantifies the uncertainty of the latent variable
and propagates it to the forecast distribution. 
GPs have been \Rev{previously} applied to smooth time series \citep{Roberts_Osborne_Ebden_Reece_Gibson_Aigrain_2013, Corani_Benavoli_Zaffalon_2021}, \Rev{but not yet} to intermittent time series.
Dealing with smooth time series, the GP \Revnew{latent function} is generally coupled with a Gaussian forecast distribution.
However, this is not suitable for intermittent time series, which require a distribution
defined over positive values and with a mass in zero.
\Revnew{Also the Poisson distribution is not a suitable forecast distribution for intermittent demand because of its short tails, which do not cover demand spikes.} 

\Revnew{A suitable distribution is, instead,}
the negative binomial \citep{Harvey_Fernandes_1989, snyder2012forecasting,Kolassa_2016}.
\Rev{Indeed,} our first model (NegBinGP) couples the GP with a negative binomial \Revnew{forecast} distribution. 
\Rev{However, this distribution is unimodal; this property might be restrictive.
Several probabilistic models} for intermittent time series have a bimodal
 distribution, with a point mass in zero and a distribution over positive values. 
 
 \Rev{In} our second model (TweedieGP) the GP \Revnew{parametrizes the mean of} a Tweedie \citep{Dunn_Smyth_2005} distribution,
which is a bimodal distribution with a mode in zero \Rev{and a long right tail}.
\Rev{TweedieGP} is the first probabilistic model for intermittent time series based on a fully parameterized Tweedie distribution.
We \Revnew{also} show (Sec.~\ref{subsec:TweedieLossDensity})
that the Tweedie loss, often used for training point forecast models on intermittent time series 
\citep{jeon2022robust, januschowski2022forecasting}, is obtained by severely approximating the Tweedie distribution.

We perform experiments on about 40'000 supply chain time series. 
Both GPs generally provide better probabilistic forecasts than the competitors.
In particular, TweedieGP often outperforms the competitors on the highest quantiles, the most difficult to estimate. This might be due to the additional flexibility of the Tweedie distribution compared to the negative binomial.
Thanks to variational methods \citep{Hensman_Matthews_Ghahramani_2015}, the training times of GPs are in line with those of other \Rev{local} models for intermittent time series.

Our implementation is based on GPyTorch \citep{Gardner_Pleiss_Weinberger_Bindel_Wilson_2018}.
Our models and the Torch implementation of the Tweedie distribution are available at the GitHub page of the project\footnote{\texttt{https://github.com/StefanoDamato/TweedieGP}}.

\section{Literature on probabilistic models for intermittent time series}
\label{sec:SoAintermittent}
\Rev{The simplest approach for forecasting intermittent time series is a static model.} This is
appropriate \citep{snyder2012forecasting} \Rev{for} time series \Rev{containing} a very large amount of zeros.
The static model can be the empirical \Rev{distribution} 
\citep[Sec 13.2]{boylan2021intermittent} or
a fitted distribution such as the negative binomial \citep{Kolassa_2016}. 

\Rev{In dynamic models, instead, it is common to decompose the intermittent time series into occurrence and demand size.}
Denoting the observation at time $t$ as $y_t$
and the indicator function as
$\mathbbm{1}_{[\cdot]}$, the occurrence and the demand size at time $t$ are:
\begin{align*}
o_t &:= \mathbbm{1}_{[ y_t > 0]}, \\
d_t & := 
\begin{cases}
 y_t \ & \text{if} \ y_t > 0 \\
 \text{undefined} &\text{otherwise}
\end{cases} \quad \Rev{,}
\end{align*} 
\Rev{respectively. Thus the observed values of a time series can be written as $y_t = o_t \cdot d_t$.}

\Rev{Several models are based on this decomposition. Croston's method \citep{Croston_1972},
predicts demand sizes and demand intervals with two independent exponential smoothing processes. The TSB method \citep{Teunter_Syntetos_Babai_2011} improves by using the second exponential smoothing
to predict the occurrence rather than the demand interval. Both \Revnew{models}
return point forecasts only.
\Revnew{Their} probabilistic counterparts have been developed by
\citet{Hyndman_2008} and \citet{snyder2012forecasting} respectively. 
The forecast distribution of both models is bimodal, \Revnew{and it is given by} a mixture of a Bernoulli and a shifted Poisson distribution.}

\Rev{\cite{snyder2012forecasting} propose a further model with a single \Revnew{latent} exponential smoothing variable that controls the mean of a negative binomial distribution; in this case the forecast distribution is 
unimodal and overdispersed.}




\citet{sbrana2023modelling} \Rev{adopts a} latent process which can be equal to zero with constant probability. The resulting forecast distribution is a mixture of a mass in zero and a truncated Gaussian distribution.

iETS \citep{Svetunkov_Boylan_2023} \Rev{is possibly the most sophisticated model in this class. It} 
assumes again the independence of occurrence and demand \Rev{size}. The latent demand \Rev{size} variable is modelled as a multiplicative exponential smoothing process.
iETS considers different models of occurrence, which are based on one or two processes; they cover cases such as demand building up, demand obsolescence, etc. 
The best occurrence model is chosen via AICc. 
The forecast distribution is the mixture of a Bernoulli and a Gamma distribution.

\Rev{The limit of the above models methods is that they do not account for the uncertainty of the
 latent processes.
Bayesian models \citep{yelland2009bayesian, chapados2014effective, Babai_Chen_Syntetos_Lengu_2021} overcome this limitation, but there is no public implementation of them, and they can be computationally intensive.
}

\Rev{WSS~\citep{Willemain_Smart_Schwarz_2004} is} a non-parametric approach based on the independence between occurrence and demand, which
 models the occurrence using a two-states Markov chain. When the simulated occurrence is positive, it samples the demand \Rev{size} from past values via bootstrapping with jittering. The forecast distribution is a mixture of a mass in zero and an integer-valued, non-parametric distribution. 
 \Rev{A limit of WSS is that it does not model the dynamic of the demand size.}

\Rev{It is} thus common to independently model occurrence and demand \Rev{size}, 
leading to a bimodal forecast distribution \Rev{with a peak mass in zero to predict the absence of demand, and a positive mode to describe demand size}.
\Rev{However, on sparse time series, the estimate of the demand size is only based on few observations, which can result in large uncertainty.}

\Rev{ADIDA \citep{Nikolopoulos_Syntetos_Boylan_Petropoulos_Assimakopoulos_2011} temporally aggregates the intermittent time series. The aggregated time series are simpler to forecast as they contain fewer zeros. The predictions on these series, made with the naïve method, are then disaggregated to the original time buckets. ADIDA can be coupled with bootstrapping or conformal inference to produce probabilistic forecasts.} 

\Rev{Combinations of point \citep{Petropoulos_Kourentzes_2015} and probabilistic forecasts \citep{Wang_Kang_Petropoulos_2024} have been used to stabilize the performance of above methods.}
\Rev{The computational cost is higher, but model selection is not required.}


\section{Gaussian Processes}
\label{sec:GP}


\Revnew{The models reviewed above often assume a specific parametric form for their latent variables. Moreover, they often estimate it with a point forecast. Gaussian Processes \citep[GPs,][]{Rasmussen_Williams_GPML} model the latent function in a non-parametric way, thus avoiding strong assumptions. Furthermore, a Gaussian Process is a Bayesian method:} parameters are treated as random variables, \Revnew{to which} a prior distribution \Revnew{is assigned}. Uncertainty is then updated based on observed data using Bayes’ theorem, resulting in a posterior distribution. The distribution of the data given the parameters is known as the likelihood \citep[Chap.~9]{Gelman_Hill_Vehtari_2020}.

A GP provides a prior distribution on the space of functions $f: \mathbb{R}_+ \rightarrow \mathbb{R}$:
\begin{equation}
\label{eq:GP1}
 f \sim \mathcal{GP}\left(m(\cdot ), k(\cdot, \cdot)\right),
\end{equation}
where $m(\cdot)$ is a mean function and $k(\cdot,\cdot)$ is a positive definite kernel.
\Rev{The prior mean encodes the knowledge we have on the latent function before observing any data. Since we have no prior information} we set $m(t)=c$ for all $t$, where $c\in \mathbb{R}$ is a learnable parameter. 

\Rev{The kernel $k$ is a positive definite function \citep[Chap.~4]{Rasmussen_Williams_GPML} and encodes the covariance between the values of $f$ at any two instants in time. We assume} $f$ to be a smooth function of time.
\Rev{This assumption is encoded by} the Radial Basis Function (RBF) kernel:
\begin{equation*}
 k(t_i, t_j) = \Rev{\sigma^2} \exp \left( - \frac{|t_i - t_j|^2}{2 \Rev{\ell}^2} \right),
 \label{eq:rbf}
\end{equation*}
\Rev{where the lengthscale $\ell \in \mathbb{R}_+$}
determines how fast the function changes in time: a smaller \Rev{$\ell$} results in quicker variations of $f$.
The outputscale $\Rev{\sigma^2} \in \mathbb{R}_+$ controls the range of values attained by the latent function.
\Rev{Both $\ell$ and $\sigma^2$ are learned via optimization; they are referred as kernel hyper-parameters.}
 

\Rev{By defining a prior in the space of functions, GPs do not need regularly spaced observations, therefore} we consider a training set of $T$ couples $(t_1, y_1), \ldots, (t_T, y_T)$, where, for each $i$, $t_i$ is the time \Rev{stamp} and $y_i$ is the observation. \Rev{We aim to generate forecast for time inputs $T+1, \dots, T+h$, where $h$ is known as forecast horizon}. 
We denote the density functions by $p(\cdot)$. The GP provides a prior on the latent vector $\mathbf{f}_{1:T} := (f(t_1), \ldots, f(t_T))^\top$ \Rev{which, given a set of time instants $t_1, \ldots,t_T$, encodes the value of the latent function at those times.} The prior is a multivariate Gaussian distribution:
\begin{equation}
\label{eq:prior}
 p(\mathbf{f}_{1:T}) = \mathcal{N}(\mathbf{f}_{1:T} \mid \mathbf{m}_{1:T}, K_{T,T}),
\end{equation}
\Rev{where $\mathbf{m}_{1:T} = (m(t_1), \ldots, m(t_T))^\top = (c, \ldots, c)^\top$ and $K_{T,T}\in \mathbb{R}^{T \times T}$ is the covariance matrix with element $K_{T,T}(i,j) = k(t_i, t_j)$.}

\begin{figure*}[!ht]
 \centering
 \includegraphics[width=1\linewidth]{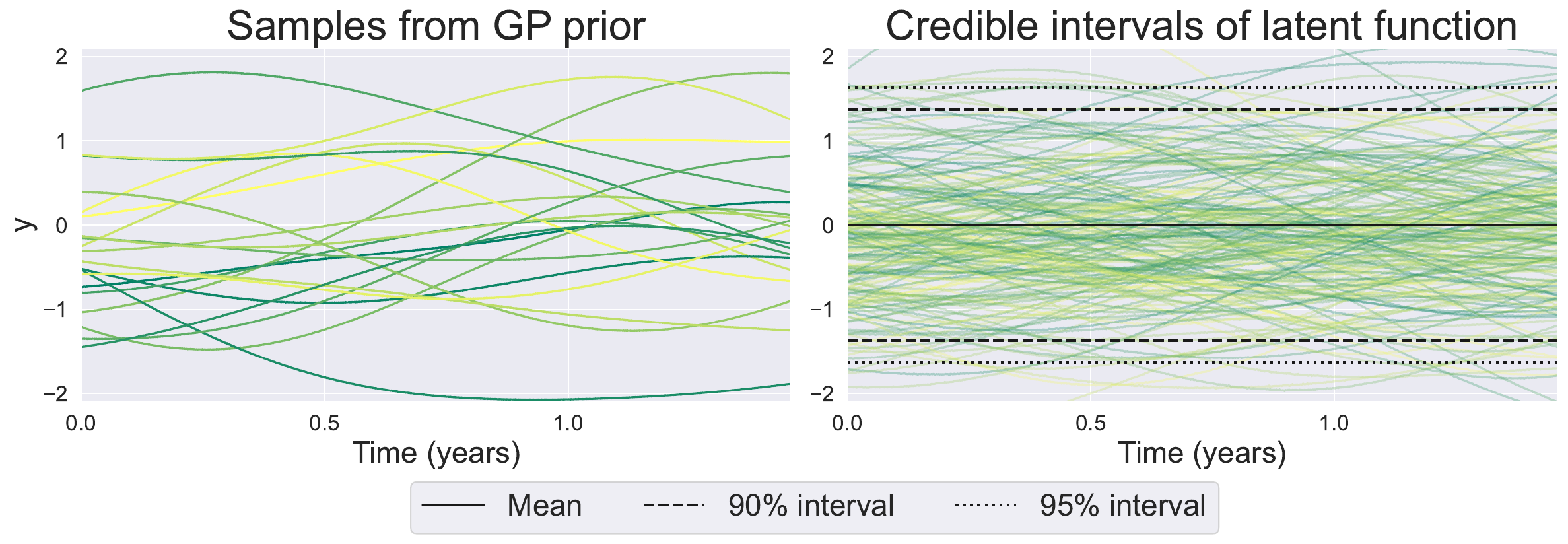}
 \caption{
 Sample trajectories \Revnew{(left panel) and credible intervals (right panel)} of the GP prior.
 The single trajectories fluctuate around the mean, while the credible intervals are flat and symmetric. Samples from $f$ can be negative.}
\label{fig:prior_samples}
\end{figure*}

Samples of vectors $\mathbf{f}_{1:T}$
are shown in Fig.~\ref{fig:prior_samples}(left) as continuous functions of $t$.
\Rev{Credible intervals display the spread of the prior distribution:} at time $t$, the $q$-level \textit{credible interval} is the interval that contains a proportion $q$ of values of $f(t)$.
A priori, the function has the same mean and variance at any time; hence credible intervals are flat (Fig.~\ref{fig:prior_samples}, right).

\begin{figure*}[!ht]
 \centering
 \includegraphics[width=1\linewidth]{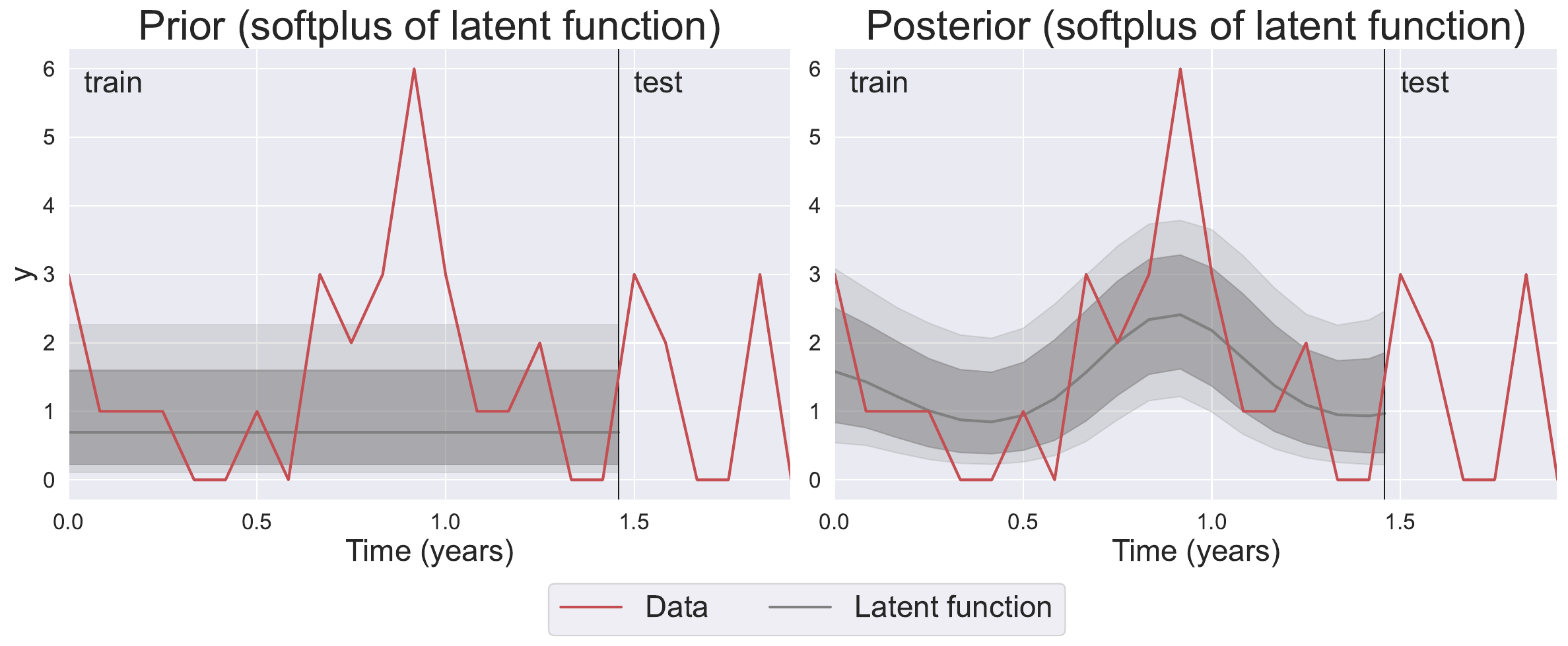}
 \caption{
 Prior (left) and posterior (right) distribution of the
 softplus of the GP latent function computed with a Tweedie likelihood. The shaded areas represent the 90\% and 95\% credible intervals. The vertical line divides train and test data. The data consist of the 1000-th time series from the Auto data set. \Rev{Fig.~\ref{fig:predicitve_forecast} (below) shows the prediction intervals for the test data.}}
 \label{fig:prior_posterior_latent}
\end{figure*}

\Rev{The latent function $f$, defined by the GP prior in eq.~\eqref{eq:GP1}, is a time-varying parameter related to the mean of the likelihood function, specified here as a negative binomial (Sec.~\ref{subsec:negBinLik}) or a Tweedie distribution (Sec.~\ref{subsec:TweedieLik}). For this reason, we require a positive latent function and we ensure this} 
by passing \Rev{$f$} through the softplus function, defined as $\softplus(x) := \log(1 + e^x)$. The softplus maps negative values of $x$ to small positive values, while it is close to the identity function for $x>2$.
A priori, the credible intervals of
$\mathrm{softplus}(f)$
are identical for all time instants, non-negative and asymmetric (Fig.~\ref{fig:prior_posterior_latent}, left).

Assuming the observations to be conditionally independent given the value of the latent function,
the likelihood function is:
\begin{equation}
\label{eq:genericLikelihood}
 p( \mathbf{y}_{1:T} \mid \mathbf{f}_{1:T}, \boldsymbol{\theta}_{\mathrm{lik}}) = \prod_{i=1}^T 
p_{\mathrm{lik}} \left(y_i \mid \mathrm{softplus}(f_i), \boldsymbol{\theta}_{\mathrm{lik}} \right),
\end{equation}
where \Rev{$f_i := f(t_i)$,} $p_{\mathrm{lik}}(\cdot \mid \mathrm{softplus}(f_i), \boldsymbol{\theta}_{\mathrm{lik}})$ is either the negative binomial or the Tweedie distribution, and $\boldsymbol{\theta}_{\mathrm{lik}}$ are their hyper-parameters,
which we discuss in Secs.~\ref{subsec:negBinLik} and~\ref{subsec:tweedie_distr}. 

After observing the training data, 
we have residual uncertainty about $\mathbf{f}_{1:T}$, described by its posterior distribution:
\begin{equation}
 p(\mathbf{f}_{1:T} \mid \mathbf{y}_{1:T}) \propto p(\mathbf{y}_{1:T} \mid \mathbf{f}_{1:T}, \boldsymbol{\theta}_{\mathrm{lik}}) p(\mathbf{f}_{1:T} ),
\label{eq:posteriorGeneric}
\end{equation}
as shown in the right panel of Fig.~\ref{fig:prior_posterior_latent}. 

The posterior in eq.~\eqref{eq:posteriorGeneric} is available in \Rev{analytical} form only for the Gaussian likelihood. 
Since our likelihoods are non-Gaussian, we approximate $p(\mathbf{f}_{1:T}~\mid~\mathbf{y}_{1:T})$ with a variational inducing point approximation \citep{Hensman_Fusi_Lawrence_2013} 
constituted by a Gaussian density, denoted by $q(\mathbf{f}_{1:T} \mid \mathbf{y}_{1:T})$, with tunable mean and covariance parameters. We follow \cite{Hensman_Matthews_Ghahramani_2015} to optimize the model parameters
as described in~\ref{sec:learning_variationalGP}. \Rev{In what follows we denote the fitted variational approximation by $q(\mathbf{f}_{1:T} \mid \mathbf{y}_{1:T})$.}

The distribution of the future values of the latent function, $\mathbf{f}_{T+1:T+h} = (f(t_{T+1}), \ldots, f(t_{T+h}))^\top$, is:
\begin{equation}
 p(\mathbf{f}_{T+1:T+h} \mid \mathbf{y}_{1:T})= \int p(\mathbf{f}_{T+1:T+h} \mid \mathbf{f}_{1:T} ) q(\mathbf{f}_{1:T} \mid \mathbf{y}_{1:T})\mathrm{d}\mathbf{f}_{1:T},
 \label{eq:pred_latent}
\end{equation}
where the integral is solved analytically since both $p$ and $q$ are Gaussian. 

\begin{figure*}[!ht]
 \centering
 \includegraphics[width=1\linewidth]{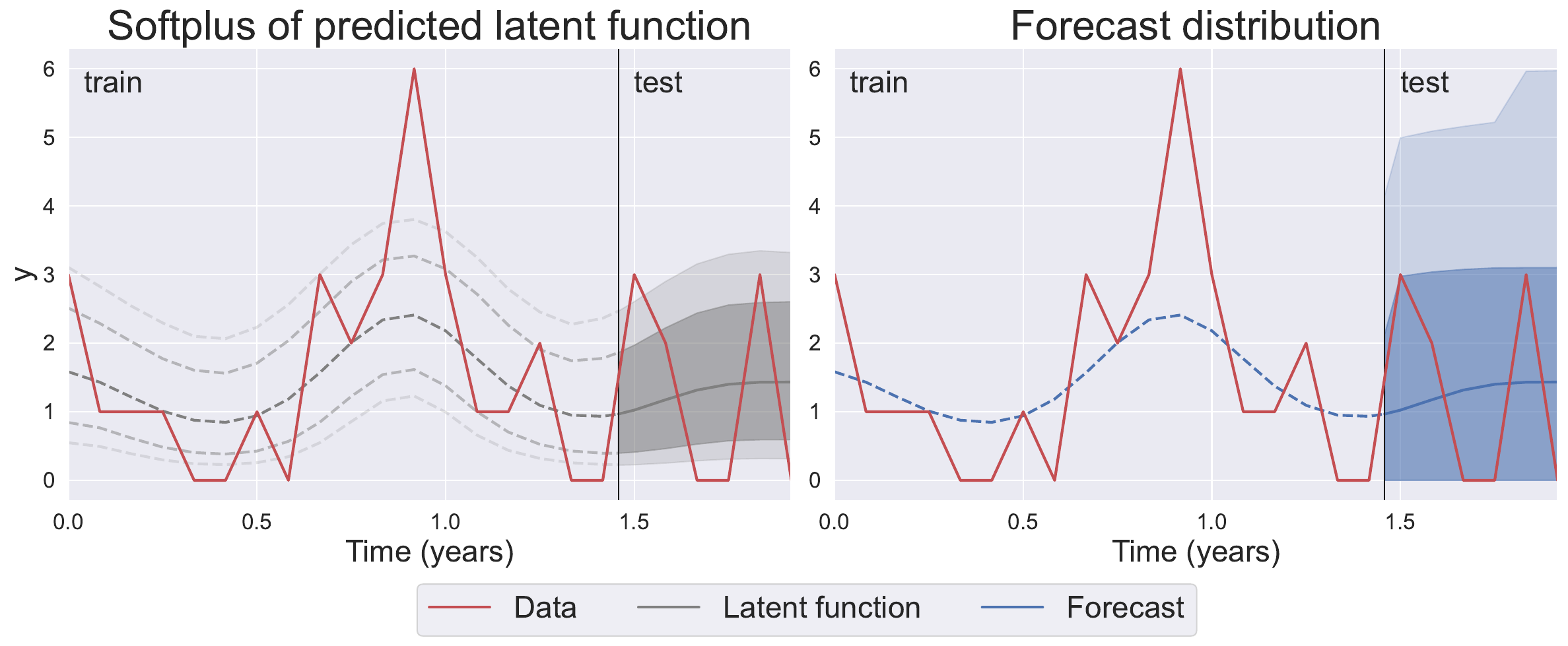}
 \caption{
 \label{fig:predicitve_forecast}
 Example of TweedieGP forecasting six steps ahead. Left: posterior distribution of the GP latent function passed through the softplus. Right: forecast distribution using the Tweedie likelihood. The shaded regions represents $90\%$ and $95\%$ prediction intervals respectively. The figures complement the model fit in Fig.~\ref{fig:prior_posterior_latent}.}
\end{figure*}

We obtain the forecast distribution 
by sampling from $p(\mathbf{f}_{T+1:T+h} \mid \mathbf{y}_{1:T})$ and passing each sample through the likelihood.
Thus the uncertainty on
$\mathbf{f}_{T+1:T+h}$
is propagated to the forecast distribution $p(\mathbf{y}_{T+1:T+h}~\mid~\mathbf{y}_{1:T})$.
Fig.~\ref{fig:predicitve_forecast} shows the posterior distribution of $\mathbf{f}_{T+1:T+h}$ (left) and the forecast distribution (right).

\Rev{The distribution of the latent function at future times, shown in eq.~\eqref{eq:pred_latent}, is obtained by marginalizing a multivariate Gaussian distribution. Given a likelihood expressed as in eq.~\eqref{eq:genericLikelihood}, $h$-steps ahead forecasts are obtained without autoregressive sampling}.
We now complete the GP model by specifying the likelihood function.

\subsection{Negative binomial likelihood}
\label{subsec:negBinLik}

\Rev{The negative binomial \Revnew{is a suitable distribution for} intermittent demand \citep{snyder2012forecasting, Kolassa_2016}. \Revnew{It addresses the limitations of the Poisson distribution,} since it can have both a large probability mass in zero and long tails.}

We denote by $\mathrm{NegBin}(y;r,\negBinp)$ the density of a negative binomial distribution\Rev{; $r$ is the} number of successes and \Rev{$\negBinp$ denotes, with a slight abuse of notation, the} success probability. The mean and the variance of the distribution linearly increase with $r$\Rev{; $\negBinp$ controls the overdispersion.} \Rev{We model $r$ through the GP, by setting} $r = \text{softplus}(f)$.
Instead, \Rev{we treat} $\negBinp$ \Rev{as a learnable} hyper-parameter ($\boldsymbol{\theta}_{\mathrm{lik}} = \{ \negBinp \}$) \Rev{which we keep} fixed for all data points of the same time series.

The likelihood of an observation $y_i$ is: 
\begin{equation*}
p_{\mathrm{lik}} \left(y_i \mid \mathrm{softplus}(f_i), \boldsymbol{\theta}_{\mathrm{lik}} \right)=
\mathrm{NegBin}(y_i ; \softplus ( f_i ), \negBinp).
\end{equation*}

\Rev{We obtain NegBinGP by plugging this likelihood in eq.~\eqref{eq:genericLikelihood} and by following the procedure in the previous section.}

\subsection{The Tweedie likelihood}
\label{subsec:TweedieLik}

The Tweedie is a family of \Rev{exponential dispersion models \citep[a generalization of the exponential family, see][]{Dunn_Smyth_2005}} characterized by a power mean-variance relationship. A non-negative random variable $Y$ is distributed as a Tweedie, $Y \sim \mathrm{Tw}(\mu, \phi, \rho)$, if \Rev{$Y$ follows an exponential dispersion model distribution such that}
\begin{equation*}
 \label{eq:mean-var}
 \mathrm{Var}[Y] = \phi \mu^\rho,
\end{equation*} 
where $\mu > 0$ is the mean, $\rho > 0$ is the \textit{power} and $\phi > 0$ is the \textit{dispersion} parameter.

\label{subsec:tweedie_distr}
\begin{figure*}[!ht]
 \centering
 \includegraphics[width=1\linewidth]{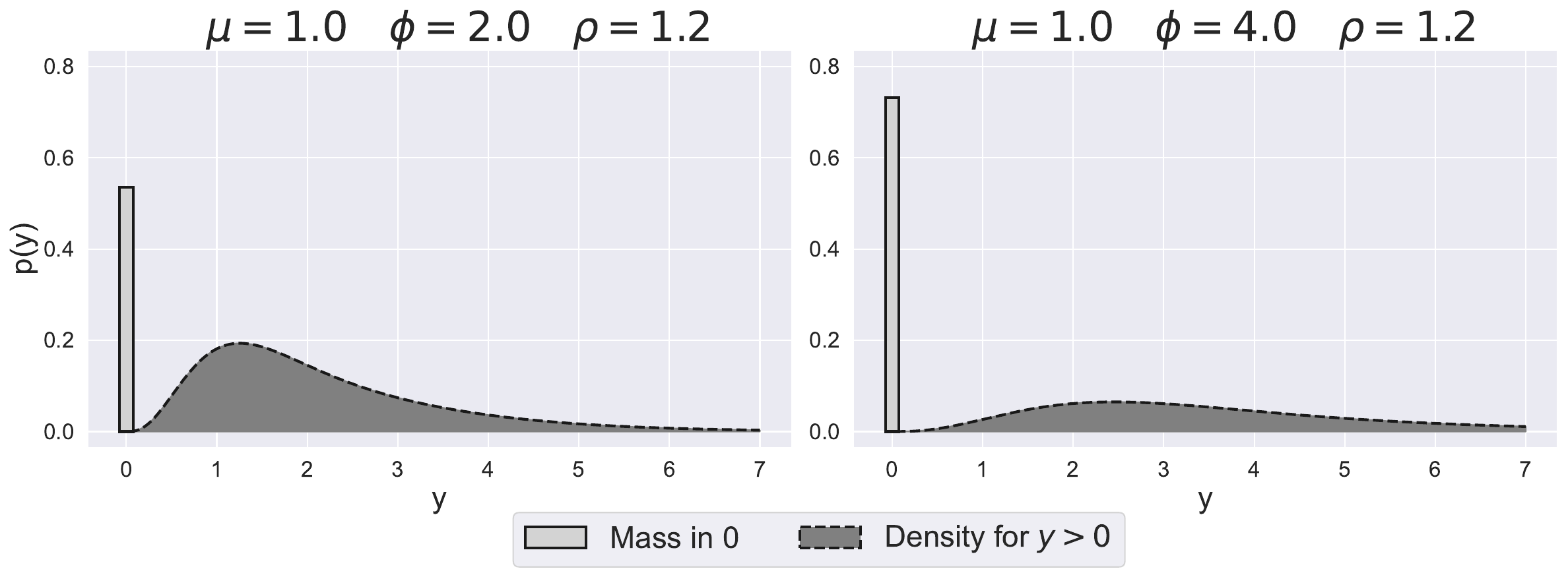}
 \caption{
 The Tweedie distribution is flexible and \Rev{possibly} bimodal. The two distributions have the same mean ($\mu=1$), but the right one has higher dispersion 
 ($\phi$)\Rev{, implying} a larger mass in 0 and in a longer right tail.}
 \label{fig:tweedie}
\end{figure*}
\Rev{Depending on $\rho$, the Tweedie family includes the Gaussian ($\rho = 0$), Poisson ($\rho =1$), and Gamma ($\rho =2$) distributions. However, we restrict $\rho \in (1,2)$ to have a non-negative distribution with a possible point mass in zero. In this setting,} the Tweedie \citep{Dunn_Smyth_2005} \Rev{is} a Poisson mixture of Gamma distributions: 
\begin{align*}
Y & \sim \sum_{i=0}^N X_i, \\
X_i & \iid \mathrm{Gamma}(\alpha, \beta),\\
N & \sim \mathrm{Poisson}(\lambda),
\end{align*}
with
\begin{equation}
\label{eq:reparam_tw}
\lambda = \frac{\mu^{2-\rho}}{\phi (2 - \rho)}, \quad \alpha = \frac{2-\rho}{\rho -1} ,\quad \beta = \frac{1}{\phi (\rho -1) \mu^{\rho -1}}. 
\end{equation}
The mass in 0 and the density for $y>0$ are:
\begin{align}
 \mathbb{P} (Y = 0) & = \mathbb{P} (N = 0) = e^{-\lambda},
 \label{eq:probZero}\\
p(y \mid \lambda, \alpha, \beta) & = \sum_{n=1}^{+\infty}
 e^{-\lambda} \frac{\lambda^n}{n!} \cdot \mathrm{Ga}(y \mid n\cdot \alpha, \beta),
 \label{eq:TwDensitySampling}
\end{align}
where \Rev{$\mathbb{P}$ denotes the probability of an event and} $\mathrm{Ga}(\cdot)$ is the density of a Gamma distribution (parametrized by shape and rate). 
The Tweedie distribution is generally bimodal (Fig.~\ref{fig:tweedie}): the first mode is in $0$, while the second one is the mode of the \Rev{mixture} of Gamma distributions. 
\Rev{This mixture has a continuous density function on the positive real values. Count forecasts can be generated by rounding the samples}.

The parametrization (\ref{eq:probZero}-\ref{eq:TwDensitySampling}), though interpretable, is not usable as a likelihood function. Indeed there is no theoretical result for truncating the infinite sum in eq.~\eqref{eq:TwDensitySampling} while controlling the exceedance probability. 
In order to evaluate the Tweedie we
use the $(\mu, \phi, \rho)$ parametrization 
 \citep{Dunn_Smyth_2005}.
The probability of zero and the continuous density for $y>0$ are:
\begin{align}
\mathbb{P}(Y = 0) & = \exp\left(-\frac{\mu^{2-\rho}}{\phi(2-\rho)}\right),
\label{eq:tweedie_p0}
\\
p(y\mid\mu, \phi, \rho) & = 
A(y) \cdot \exp \left[ \frac1\phi \left( y \frac{\mu^{1-\rho}}{1-\rho} - \frac{\mu^{2-\rho}}{2-\rho} \right) \right],
\label{eq:tweedie_lik}
\end{align}
with 
\begin{align}
\label{eq:A}
A(y) &= \frac{1}{y}\sum_{j=1}^{\infty} \frac{y^{j\alpha}(\rho-1)^{-j\alpha }}{\phi^{j(1+\alpha)}(2-\rho)^jj!\Gamma(j\alpha)} \\ \nonumber
&= \frac{1}{y}\sum_{j=1}^{\infty} V(j),
\end{align}
where $\alpha$ is defined in eq.~\eqref{eq:reparam_tw}. 
\cite{Dunn_Smyth_2005} provide
a truncating rule for evaluating the infinite summation of eq.~\eqref{eq:A}. It exploits the fact that $V(j)$ is a concave function of $j$. \Rev{In ~\ref{sec:truncating_sum} we report the derivations of the truncating rule and we show that we generally require less than 10 terms in the summation of eq.~\eqref{eq:A} for a precise approximation.} 
The Tweedie \Rev{density} is \Rev{complex} to implement; yet, its evaluation only requires a small additional overhead compared to other distributions.

The likelihood of TweedieGP is thus obtained by substituting in eq.~\eqref{eq:genericLikelihood}:
\begin{equation*}
p_{\mathrm{lik}} \left(y_i \mid \mathrm{softplus}(f_i), \boldsymbol{\theta}_{\mathrm{lik}} \right)=
\mathrm{Tw}(y_i ; \softplus ( f_i ), \phi, \rho), 
\end{equation*}
where $\boldsymbol{\theta}_{\mathrm{lik}} = \{\phi,\rho\}$. 
We set $\mu = \text{softplus}(f)$: the latent variable affects, through $\lambda$ and $\beta$, both the mass in $0$ and the distribution on the positive $y$, see
eq.~\eqref{eq:reparam_tw}, \eqref{eq:probZero}, 
\eqref{eq:TwDensitySampling}. Thus, a bimodal forecast distribution \Revnew{is parametrized by} a single latent process.
The hyper-parameters $\phi$ and $\rho$ are optimized and they are equal for all the data points in the same time series. 

We now discuss the relation between the Tweedie
distribution and the Tweedie loss, often
used to train point forecast models on
intermittent time series.

\subsubsection{Comparison with Tweedie loss}
\label{subsec:TweedieLossDensity}
\cite{januschowski2022forecasting} mention the \Revnew{use of the} \textit{Tweedie loss} for tree-based models as one of the reasons of \Rev{success} in the M5 competition.
Also \cite{jeon2022robust} obtained good results in the M5 competition by training \Rev{a} DeepAR \citep{salinas2020deepar} model with the Tweedie loss.
The Tweedie loss is indeed available both in \texttt{lightGBM}\footnote{\label{footnote:lightGBM}\url{https://lightgbm.readthedocs.io/en/latest/Parameters.html}}
and in \texttt{PyTorch Forecasting}\footnote{\label{footnote:TweedieLoss}\url{https://pytorch-forecasting.readthedocs.io/en/stable/api/pytorch_forecasting.metrics.point.TweedieLoss.html}}.
But such models only return point forecasts.
\Rev{We clarify, for the first time, that the Tweedie loss is obtained from the Tweedie distribution via a rough approximation, which makes it unsuitable for probabilistic forecasting.} 

The Tweedie loss \citep{jeon2022robust} is:
\begin{equation}
\mathcal{L}(\mu, \rho\mid y) =- y \frac{\mu^{1-\rho}}{1-\rho} + \frac{\mu^{2-\rho}}{2-\rho}.
\label{eq:tw_loss_l}
\end{equation}
Interpreting eq.~\eqref{eq:tw_loss_l} as a negative log-likelihood, the implied likelihood is:
\begin{equation}
 p (y \mid \mu, \rho) = \exp \left( y \frac{\mu^{1-\rho}}{1-\rho} - \frac{\mu^{2-\rho}}{2-\rho} \right),
 \label{eq:tw_loss_p}
\end{equation}
which approximates the Tweedie density in eq.~\eqref{eq:tweedie_lik} by setting both $A\Rev{(y)}=1$ and $\phi=1$.
Setting $A\Rev{(y)}=1$ \Rev{for all $y>0$} avoids \Rev{evaluating} the infinite summation in eq.~\eqref{eq:A}; however, this makes the distribution unimodal and shortens its tails.
Indeed, $A$ is a function of $y$ and not a simple normalization constant.
Moreover, fixing the dispersion parameter to $\phi=1$ further reduces the flexibility.
In Sec.~\ref{subsec:likComparison} we 
show that the accuracy of TweedieGP
worsens when we \Rev{adopt} the approximation with $A\Rev{(y)}=1$ and $\phi=1$,
rather than the actual Tweedie distribution.

We can nevertheless understand why
the function in eq.~\eqref{eq:tw_loss_l} is an effective loss function for point forecast on intermittent time series. The first term of the exponential corresponds to an unnormalised exponential distribution and the second term is a penalty term which forces $\mu$ to remain close to zero.
We illustrate the shape of the unnormalized distribution from eq.~\eqref{eq:tw_loss_p} in~\ref{sec:comparison_tweedie_loss}.

\section{Experiments} \label{sec:experiments}

\begin{figure*}[!ht]
 \centering
\includegraphics[width=1\linewidth]{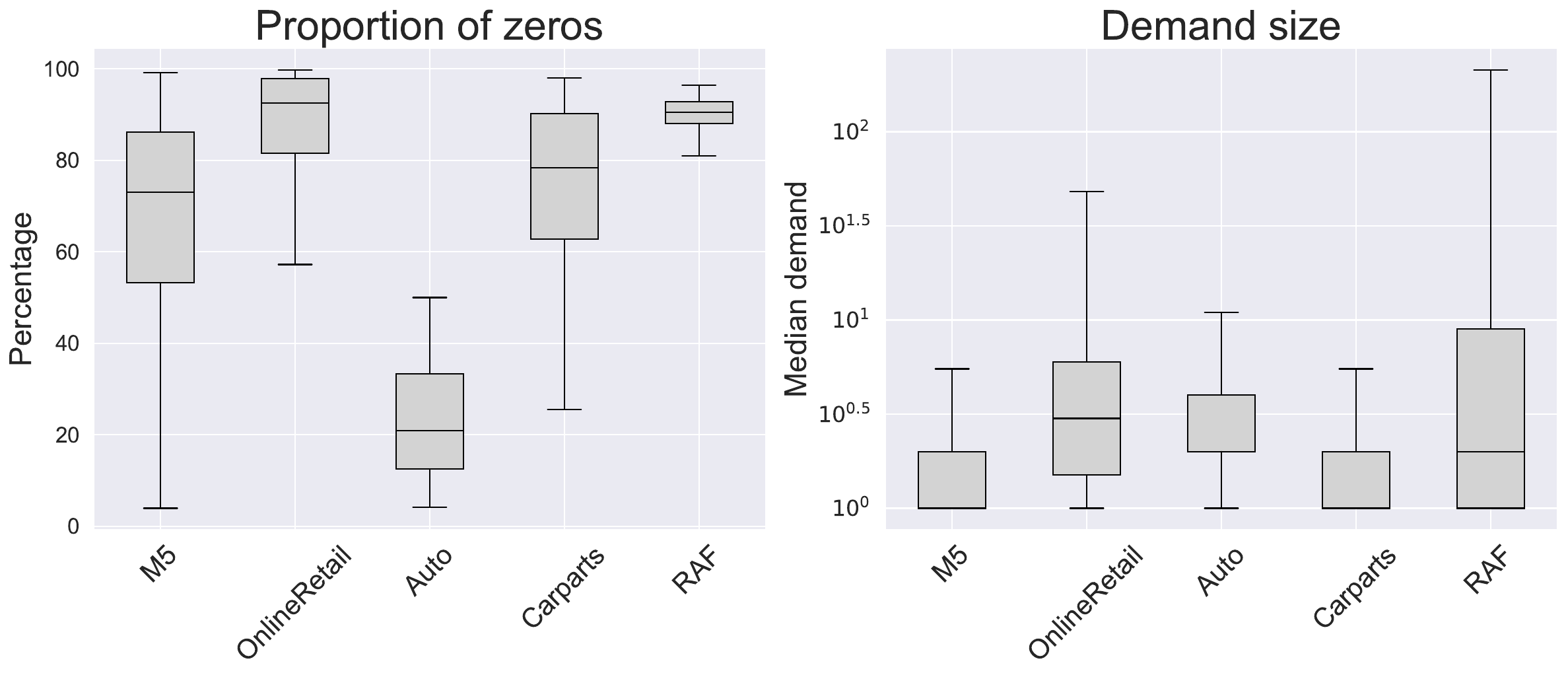}
 \caption{Proportion of zeros and \Rev{median} demand \Rev{size} in the different data sets; the y-axis of the median demand \Rev{size} is in \Rev{$\log_{10}$} scale.}
 \label{fig:datasets}
 
\end{figure*}
\begin{table}[!h] 
\centering
\begin{tabular}{l|rcrr}
\toprule
Name & \# of t.s. & Freq & $T$ & $h$ \\ 
\midrule
\rowcolor{backcolour}
M5 & 30490 & Daily & 1941 & 28 \\ 
OnlineRetail & 2032 & Daily & 346 & 28 \\
\rowcolor{backcolour}
Auto & 3000 & Monthly & 18 & 6 \\ 
Carparts & 2503 & Monthly & 45 & 6 \\ 
\rowcolor{backcolour}
RAF & 5000 & Monthly & 72 & 12 \\ 
\bottomrule
\end{tabular}
\caption{Characteristics of the extracted intermittent time series.}
\label{tab:datasets}
\end{table}

\Revnew{We test the performance of our models on 5 publicly available data sets. We consider a time series as intermittent if the average demand interval (ADI) is strictly greater than 1, that is, all the time series containing at least one observation equal to zero \citep{Syntetos_Boylan_Croston_2005}. The data sets are:}

\Revnew{\begin{itemize}
\item M5\Rev{\footnote{\url{https://www.kaggle.com/competitions/m5-forecasting-accuracy/overview}}}: daily sales data from some US Walmart stores, released for a popular competition \citep{makridakis2022m5}. 
\item OnlineRetail\Rev{\footnote{\url{https://archive.ics.uci.edu/dataset/352/online+retail}}}: daily sales records from a British online store. For each item, we obtained a time series aggregating the daily sales. We only retained time series showing positive demand in one of the first 200 days.
\item Auto\Rev{\footnote{\url{https://github.com/canerturkmen/gluon-ts/tree/intermittent-datasets/datasets/intermittent_auto}}}: short monthly time series of automobile parts, weakly intermittent. 
\item Carparts\Rev{\footnote{\url{https://zenodo.org/records/4656022#.YYPvWMYo-Ak}}}: longer monthly time series on spare parts for cars, strongly intermittent. 
\item RAF\Rev{\footnote{\url{https://github.com/canerturkmen/gluon-ts/tree/intermittent-datasets/datasets/intermittent_raf}}}: 
monthly time series of spare parts for British Royal Air Force aircrafts.
\end{itemize}}


In Tab.~\ref{tab:datasets} we report \Rev{the length of the training data ($T$) and of the forecast horizon ($h$) of each data set; \Revnew{e consider overall} more than 40'000 intermittent time series.}
Fig.~\ref{fig:datasets} shows how the proportion of zeros and the demand size varies across the time series of each data set.

The data sets with the highest proportion of zeros are RAF and OnlineRetail (Fig.~\ref{fig:datasets} left). \Rev{However,} the time series of OnlineRetail are about five times longer than those of RAF, Tab.~\ref{tab:datasets}.
In contrast, Auto has the lowest proportion of zeros and its time series are also the shortest ($T=18$).
The M5 data set is the most heterogeneous as for proportion of zeros and it generally has narrow demand \Rev{size}. It also contains the largest amount (30490) of time series, which are also very long ($T=1941$). 

\subsection{GP training}\label{sec:gptraining}

We implement our models in GPyTorch \citep{Gardner_Pleiss_Weinberger_Bindel_Wilson_2018}.
We estimate the variational parameters, the hyper-parameters of the GP ($c, \Rev{\ell},\sigma^2$) and of the likelihood ($\negBinp$ for the negative binomial distribution, $\phi$ and $\rho$ for the Tweedie) via gradient descent for 100 iterations with early stopping, using Adam optimizer \citep{Kingma_Ba_2017} with learning rate $0.1$.

We train a GP for each time series.
For both NegBinGP and TweedieGP, we obtain
the forecast distribution by drawing 50'000 samples.
Training \Rev{and prediction} times are \Rev{of the same order of magnitude of iETS},
as detailed in Sec.~\ref{sec:computational_times}. 
Rarely, the training \Rev{fails} due to numerical issues; when this happens, we restart the optimization. 

Before training TweedieGP, \Rev{we scale the observations by dividing them by the median demand size, that is, we divide each value by $\mathrm{median}(\{y_i : y_i > 0,\ i = 1, \dots, T\})$}; this simplifies the optimization.
This scaling can be applied as the zeroes remain unchanged and the Tweedie \Rev{distribution} is absolutely continuous on positive data.
We bring the samples back to the original scale after drawing them.
We evaluate the effect of the scaling in Sec.~\ref{subsec:likComparison}.

\subsection{Baselines}
We compare NegBinGP and TweedieGP against empirical quantiles (EmpQuant), WSS, \ADIDA, and iETS. \Rev{We implemented on our own
WSS \citep{Willemain_Smart_Schwarz_2004}, drawing 50'000 samples from its forecast distribution.} 

We \Rev{use}
ADIDA \citep{Nikolopoulos_Syntetos_Boylan_Petropoulos_Assimakopoulos_2011}
\Rev{generating prediction intervals} via conformal inference \citep{Angelopoulos_Bates_2022}, 
\Rev{using} the implementation provided \Rev{by} \texttt{statsforecast} (\citet{Garza_Canseco_Challú_Olivares_2022}, v.~1.7.6).
We refer \Rev{to it as} \ADIDA, \Rev{where the subscript stands for conformal}.

We use iETS \citep{Svetunkov_Boylan_2023} from the \texttt{smooth} R package (\citet{Svetunkov_smooth}, v.~4.1.1).
We use the most complete model (iETS$_A$) which fits four different occurrence models and selects among them via AICc. \Rev{We draw 50'000 samples from the forecast distribution.}

In~\ref{sec:reproducibility}, we show how to reproduce our experiments and we detail the integration of our models in GPyTorch \citep{Gardner_Pleiss_Weinberger_Bindel_Wilson_2018}.

\Revnew{We do not include global models such as LightGBM \citep{lighGBM}, which performed well on the M5 competition \citep{januschowski2022forecasting}, since we focus our comparison on local models. We leave to future works the comparison between local and global models.}

\subsection{Metrics}

Denoting by $\hat{y}^q$ the forecast of quantile $q$, and by $y$ the observed value, the quantile loss \citep{Gneiting_Raftery_2007} is:
\begin{equation}
 \mathrm{Q}_{q}(\hat{y}^q, y) = 
2 \cdot\begin{cases} q (y -\hat{y}^q) \quad& \text{if} \ y \geq \hat{y}^q \\ 
(1-q)(\hat{y}^q - y) \quad& \text{else} \end{cases}.
 \label{eq:quantileLoss}
\end{equation}

For high \Rev{quantiles} $q$, this metric strongly penalises cases in which $y$ exceeds the predicted value $\hat{y}^q$, that is, when the demand \Rev{size} is underestimated. We evaluate $\mathrm{Q}_{q}(\hat{y}^q, y)$ for the quantile levels $q \in \{ 0.5, 0.8, 0.9, 0.95, 0.99 \}$. %
We do not assess quantiles lower than $0.5$ because they are generally zero for all models. 

The quantile loss is however scale-dependent. In order to obtain a scale-free indicator \Rev{\citep{Athanasopoulos_Kourentzes_2023}} we scale it by the quantile loss of the empirical quantiles ($\mathrm{emp}^q$) on the training data, \Rev{and we average it through the forecast horizon. Let $\hat{\mathbf{y}}^q_{T+1:T+h} := (\hat{y}_{T+1}^q, \dots, \hat{y}_{T+h}^q)^\top$ be the forecasts of quantile $q$ from times $T+1$ to $T+h$; the scaled quantile loss is:
\begin{equation*}
 \mathrm{sQ}_{q}(\hat{\mathbf{y}}^q_{T+1:T+h}, \mathbf{y}_{T+1:T+h}) = \frac{ \frac1h \sum_{t=1}^{h}\mathrm{Q}_{q}(\hat{y}^q_{T+t}, y_{T+t})} {\frac1T \sum_{t=1}^T \mathrm{Q}_{q}(\mathrm{emp}^q, y_t)}. 
\end{equation*}
}
The values \Rev{of $\mathrm{sQ}_q$} are usually greater than 1 since the scaling factor, based on the training set, is optimistically biased. 

\Rev{
The scaled quantile loss scores the quantile predictions as point forecasts. We then assess the entire predictive distribution, using an indicator similar to the Ranked Probability Score \citep[RPS,][]{Czado_Gneiting_Held_2009}}.
\Rev{The RPS can be approximated by averaging the quantile loss computed for multiple probability levels \citep{Bracher_Ray_Gneiting_Reich_2021}.
For many applications involving intermittent time series, the lower part of the forecast distribution is not relevant \citep{boylan_syntetos_2006}. We thus focus on a set of probability levels $L = \{0.5, 0.55, 0.6, 0.65, 0.7, 0.75, 0.8, 0.85, 0.9, 0.95, 0.99\}$ with cardinality \Revnew{$|L|=11$} and we propose the metric
\[
 \CRPS (\{\hat{y}^q\}^{q \in L}, y) = \frac{1}{|L|} \sum_{q \in L} \mathrm{Q}_{q}(\hat{y}^q, y),
\]
where the subscript shows that we only consider quantiles equal to or greater than 0.5. We make it scale-independent by scaling it 
by $\CRPS$ of the empirical quantiles:
\begin{equation*}
 \mathrm{s}\CRPS (\{\hat{\mathbf{y}}_{T+1:T+h}^q \}^{q \in L}, \mathbf{y}_{T+1:T+h} ) = \frac{\frac1h \sum_{t=1}^h \CRPS (\{\hat{y}_{T+t}^q\}^{q \in L}, y_{T+t})}{\frac1T \sum_{t=1}^T \CRPS (\{\mathrm{emp}^q\}^{q \in L}, y_t )}.
\end{equation*}
}

\Rev{
We also report the Root Mean Squared Scaled Error (RMSSE) \citep{Hyndman_Koehler_2006} of the point forecasts:
\[
\RMSSE(\hat{\mathbf{y}}_{T+1:T+h}, \mathbf{y}_{T+1:T+h}) = \sqrt{\frac{\frac1h \sum_{t=1}^h \left(\hat{y}_{T+t} - y_{T+t}\right)^2} {\frac{1}{T-1} \sum_{t=2}^T \left(y_{t} - y_{t-1} \right)^2}} 
\]
where $\hat{\mathbf{y}}_{T+1:T+h} := (\hat{y}_{T+1}, \dots, \hat{y}_{T+h})^\top$ denotes a vector of forecasts.
We use the mean of the forecast distribution as point forecast, as this is the minimizer of RMSSE \citep{kolassa2020best}.}

Finally, we measure the forecast coverage, i.e., the proportion of observations that lie within the $q$-level prediction interval.
An ideal forecast has coverage $q$.
\Rev{However, for discrete data 
concentrated around low values, 
actual quantiles exist 
only at a coarse probability level;
for instance we might have
$\hat{y}^{0.2} = \hat{y}^{0.5} = 0$.} 

\subsection{Discussion}

\begin{table}[!ht]
 \centering
 \begin{tabular}{ll|cccccc}
 \toprule
 \rotatebox[origin=c]{90}{Data set} & \rotatebox[origin=c]{90}{Metric} & \rotatebox[origin=c]{90}{EmpQuant} & \rotatebox[origin=c]{90}{WSS} & \rotatebox[origin=c]{90}{\ADIDA} & \rotatebox[origin=c]{90}{iETS} & \rotatebox[origin=c]{90}{NegBinGP} & \rotatebox[origin=c]{90}{TweedieGP} \\ 
 \midrule 
 \greytabline & $\mathrm{sQ}_{0.5}$ & 1.84 & 1.83 & 1.83 & 1.78 & \textbf{1.74} & 1.75 \\
 & $\mathrm{sQ}_{0.8}$ & 1.71 & 1.74 & \textbf{1.43} & 1.61 & 1.46 & 1.45 \\
\greytabline & $\mathrm{sQ}_{0.9}$ & 1.54 & 1.62 & 1.33 & 1.53 & 1.28 & \textbf{1.26} \\
& $\mathrm{sQ}_{0.95}$ & 1.38 & 1.49 & 1.34 & 1.58 & 1.18 & \textbf{1.16} \\
\greytabline & $\mathrm{sQ}_{0.99}$ & 1.24 & 1.39 & 1.65 & 2.20 & \textbf{1.16} & \textbf{1.16} \\
 & s\CRPS & 1.70 & 1.71 & 1.54 & 1.67 & 1.48 & \textbf{1.47} \\
\greytabline \multirow[c]{-7}{*}{\rotatebox[origin=c]{90}{M5}}
 & \RMSSE & 0.98 & 0.99 & \textbf{0.93} & \textbf{0.93} & 0.94 & \textbf{0.93} \\
\midrule 

 & $\mathrm{sQ}_{0.5}$ & 2.35 & 2.35 & 2.54 & 2.35 & \textbf{2.33} & 2.34 \\
\greytabline & $\mathrm{sQ}_{0.8}$ & 2.33 & 2.33 & 2.28 & 2.39 & \textbf{2.24} &\textbf{2.25} \\
 & $\mathrm{sQ}_{0.9}$ & 2.31 & 2.29 & 2.23 & 2.50 & \textbf{2.19} & \textbf{2.19} \\ 
\greytabline & $\mathrm{sQ}_{0.95}$ & 2.33 & 2.29 & 2.34 & 2.75 & 2.24 & \textbf{2.21} \\
 & $\mathrm{sQ}_{0.99}$ & 3.01 & \textbf{2.88} & 4.08 & 4.35 & 3.09 & 2.98 \\
\greytabline & s\CRPS & 2.36 & 2.35 & 2.44 & 2.50 & \textbf{2.29} & \textbf{2.29} \\
 \multirow[c]{-7}{*}{\rotatebox[origin=c]{90}{OnlineRetail}} 
& \RMSSE & 0.97 & 0.98 & \textbf{0.94} & \textbf{0.94} & \textbf{0.94} & 0.95 \\
\midrule 

\greytabline & $\mathrm{sQ}_{0.5}$ & 1.19 & 1.35 & 1.25 & 1.20 & \textbf{1.17} & 1.19 \\
 & $\mathrm{sQ}_{0.8}$ & 1.31 & 1.47 & 1.36 & 1.36 & \textbf{1.30} & \textbf{1.30} \\
\greytabline & $\mathrm{sQ}_{0.9}$ & 1.48 & 1.58 & 1.58 & 1.50 & \textbf{1.42} & \textbf{1.42} \\
 & $\mathrm{sQ}_{0.95}$ & 1.80 & 1.77 & 2.17 & 1.71 & \textbf{1.62} & \textbf{1.63} \\
\greytabline & $\mathrm{sQ}_{0.99}$ & 4.16 & 2.55 & 7.21 & 2.85 & \textbf{2.43} & 2.63 \\
 & s\CRPS & 1.29 & 1.44 & 1.40 & 1.31 & \textbf{1.27} & \textbf{1.27} \\
\greytabline \multirow[c]{-7}{*}{\rotatebox[origin=c]{90}{Auto}}
& \RMSSE & \textbf{0.82} & 0.93 & 0.84 & \textbf{0.82} & \textbf{0.82} & \textbf{0.82} \\
\midrule 

 & $\mathrm{sQ}_{0.5}$ & 1.13 & 1.15 & 1.20 & \textbf{1.10} & \textbf{1.10} & 1.11 \\
\greytabline & $\mathrm{sQ}_{0.8}$ & 1.18 & 1.33 & 1.14 & 1.15 & \textbf{1.09} & \textbf{1.09} \\
 & $\mathrm{sQ}_{0.9}$ & 1.25 & 1.45 & 1.16 & 1.24 & 1.15 & \textbf{1.13} \\
\greytabline & $\mathrm{sQ}_{0.95}$ & 1.32 & 1.63 & 1.34 & 1.46 & 1.23 & \textbf{1.20} \\
 & $\mathrm{sQ}_{0.99}$ & 1.86 & 2.00 & 3.71 & 3.03 & 1.64 & \textbf{1.55} \\
\greytabline & s\CRPS & 1.19 & 1.29 & 1.25 & 1.18 & \textbf{1.11} & \textbf{1.10} \\
 \multirow[c]{-7}{*}{\rotatebox[origin=c]{90}{Carparts}} 
& \RMSSE & 0.66 & 0.78 & 0.60 & \textbf{0.59} & 0.61 & 0.61 \\
 \midrule 
 
\greytabline & $\mathrm{sQ}_{0.5}$ & \textbf{1.00} & \textbf{1.00} & 1.37 & \textbf{1.00} & \textbf{1.00} & \textbf{1.00} \\
 & $\mathrm{sQ}_{0.8}$ & \textbf{1.00} & 1.01 & 1.22 & \textbf{1.00} & 1.01 & 1.02 \\
\greytabline & $\mathrm{sQ}_{0.9}$ & 1.10 & 1.16 & 1.17 & \textbf{1.07} & 1.09 & 1.15 \\
 & $\mathrm{sQ}_{0.95}$ & \textbf{1.24} & 1.43 & 1.35 & 1.38 & \textbf{1.24} & 1.26 \\
\greytabline & $\mathrm{sQ}_{0.99}$ & 2.12 & 2.21 & 3.79 & 3.89 & 2.16 & \textbf{2.09} \\
 & s\CRPS & \textbf{1.06} & 1.09 & 1.38 & 1.12 & \textbf{1.06} & 1.08 \\
\greytabline \multirow[c]{-7}{*}{\rotatebox[origin=c]{90}{RAF}} 
& \RMSSE & 0.61 & 0.65 & 0.61 & \textbf{0.59} & 0.60 & 0.60 \\
 \bottomrule
\end{tabular}
 \caption{\Rev{Metrics averaged over the time series of each data set.}}
 \label{tab:s_bs}
\end{table}

We show in Tab.~\ref{tab:s_bs} the score of each method, \Rev{averaged with respect to all the time series of each data set.}
In each row, we boldface the model with the lowest average loss and the models whose average loss is not significantly different from the best. We test significance of the difference in loss by using the paired $t$-test with FDR correction for multiple comparisons \citep[Sec 18.7.1]{hastie2009elements}. 

\Rev{We argue that \Revnew{models should be compared considering} the mean of the quantile losses, and not a rank-based statistic. The distribution of the quantile loss is strongly asymmetric: most forecasts have a small quantile loss; only a few forecasts incur a large \Revnew{penalty}.
The mean is sensitive to such rare and large losses, unlike rank statistics.
Moreover, the correct predictive distribution is guaranteed to minimize the expected value of the scoring rule \citep{Gneiting_Raftery_2007}, but there are no guarantees for rank statistics.
}



\paragraph{Scores} 
TweedieGP is \Rev{often} the best performing model on the highest quantiles $\{0.9, 0.95,
0.99\}$, the most important for decision making. \Rev{This} might be due to Tweedie distribution’s ability to extend its tails.

TweedieGP and NegBinGP perform similarly on data sets containing short time series (Auto, Carparts, RAF)\Rev{, where} the effect of the GP prior might be more important than the choice of the likelihood.
In contrast, on data sets containing longer time series
(OnlineRetail, M5), TweedieGP has generally an advantage over NegBinGP. 
The advantage of TweedieGP on the highest quantiles is emphasized on OnlineRetail, whose time series are both long and with high demand \Rev{sizes} (see Fig.~\ref{fig:datasets}), challenging the long tail of the forecast distribution.

iETS is a good match for the GP models \Rev{on the lowest quantile we consider}, but it is generally outperformed on the highest quantiles. \Rev{Also in \citet{Svetunkov_Boylan_2023},
iETS is outperformed by other methods on 
the average quantile loss.}

The empirical quantiles are generally preferable to both WSS and \ADIDA. 
In particular on the RAF data set, which is characterized by the largest amount of zeros, the empirical quantiles match the performance of the GPs.
This confirms the suitability of static models for very sparse time series; it also shows that the GP model adapts well to different types of temporal correlations, becoming an i.i.d. model when needed.
\Rev{The results of the approximate RPS are in line with those of the individual quantiles:} \Revnew{TweedieGP is superior on long time series, and is comparable to NegBinGP on shorter series. Both methods outperform the competitors on all datasets except RAF.}

\Rev{The best models according to RMSSE are iETS, \ADIDA, and GPs. The differences observed between these methods are often very small. However, \ADIDA~performs poorly on the quantile loss, showing
that conformal inference is currently unsuitable for probabilistic forecasting on count data.} 

\paragraph{ADI sensitivity}
\Revnew{The model ranking remains consistent if we consider as intermittent only the time series with ADI $> 1.32$. This cut-off was introduced by \citet{Syntetos_Boylan_Croston_2005}, studying whether Croston's method or the Syntetos-Boylan approximation \citep{Syntetos_Boylan_2005} yields a lower MSE. It is also sometimes used as a cut-off for intermittent demand \citep{Long_Bui_Oktavian_Schmidt_Bergmeir_Godahewa_Lee_Zhao_Condylis_2025}; the practical effect is that the selected time series are more sparse. In the M5, OnlineRetail, Carparts, and RAF data sets,
this criterion reduced the number of time series by less than 5\%, implying only minor changes to the numerical results.
The only exception is Auto, where only 1227 time series out of 3000 are retained. Also in this case, however, the model rankings do not change, providing favorable evidence for the GP models. The results are reported in~\ref{sec:res_1.32}.}

\begin{figure*}[!htp]
 \centering
\includegraphics[width=1\linewidth]{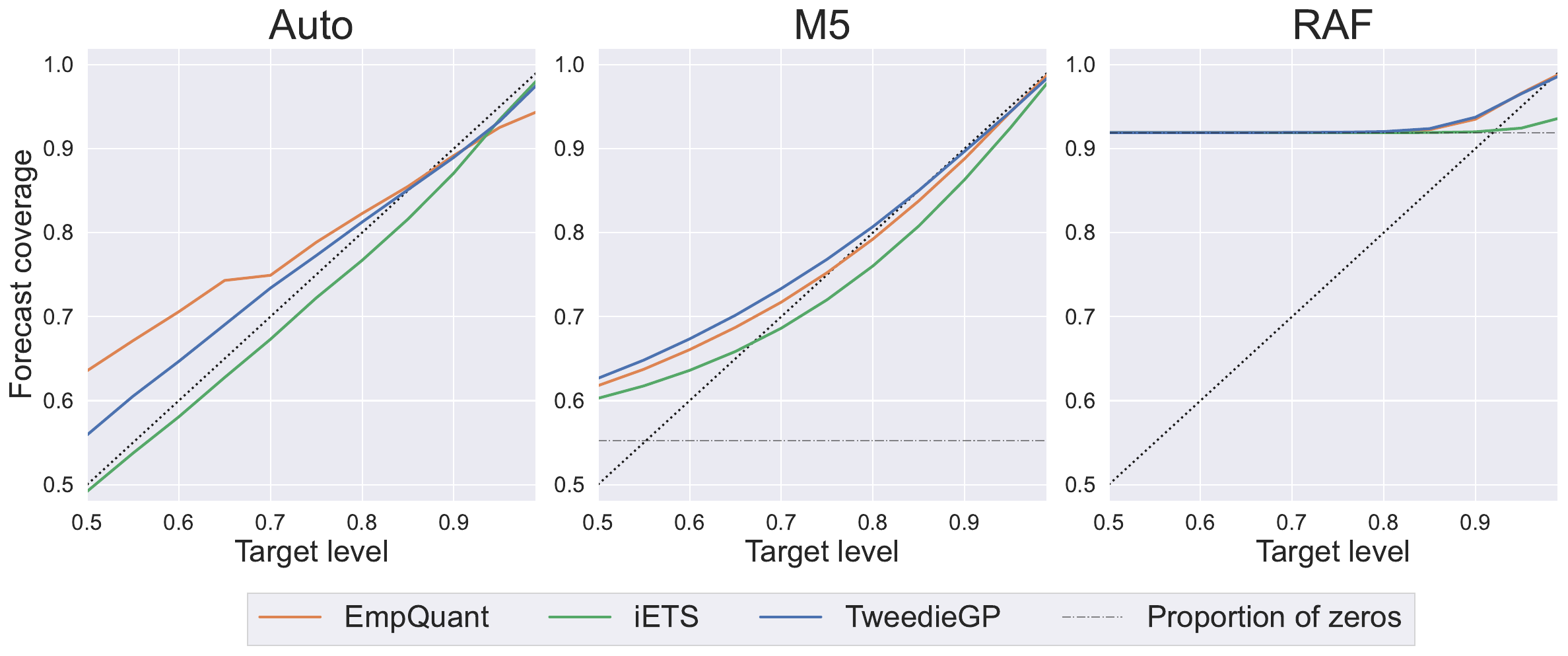}
 \caption{The coverage cannot go below the proportion of zeros in the test set, shown by
 a dashed horizontal line. On RAF, 
 this implies overcoverage of most quantiles, apart from the highest ones.}
 \label{fig:calibrations}
\end{figure*}

\paragraph{Coverage}Fig.~\ref{fig:calibrations} shows the \Revnew{coverage} of selected methods
(empirical quantiles, iETS and TweedieGP) on three data sets (Auto, M5 and RAF) \Rev{containing respectively the shortest, longest and most sparse time series}.
Better forecasts lie closer to the dotted diagonal line.
On intermittent time series, the coverage is at least as large as the proportion of zeros; there is hence
overcoverage of the lower quantiles, as clear from the third panel of Fig.~\ref{fig:calibrations}.
Indeed, for a discrete variable that is concentrated around low values, meaningful quantiles only exist for high enough probability level; e.g., \Rev{a forecast of quantile 0.5 may in fact be a forecast of quantile 0.8.}

On the Auto data set, which has moderate intermittency, TweedieGP \Rev{has} almost correct coverage on all quantiles above $0.5$. 
A similar comment can be done for M5, even though in this case the proportion of zeros is larger. Correct coverage is provided, only by TweedieGP and empirical quantiles, for quantiles above $0.8$. 
On the RAF data set, the very sporadic \Rev{positive} demand implies forecast overcoverage apart from the highest quantiles.

It is also interesting to analyze how the models compare on the quantile loss when they are correctly calibrated.
For instance, empirical quantiles, iETS and TweedieGP provide correct coverage on the 95-th \Rev{percentile} of M5.
On Auto, both TweedieGP and the empirical quantiles provide correct coverage on \Rev{quantile 0.9}.
In these cases, TweedieGP provides the lowest quantile loss (Tab.~\ref{tab:s_bs}) among correctly 
calibrated models. 
Recall that the quantile loss combines a reward for the sharpness and a miscoverage penalty.
Hence, given the same calibration, TweedieGP achieves lower loss as it provides a sharper estimate of the quantiles. 

\paragraph{Computational times}\label{sec:computational_times}

\begin{table*}[!h]
\centering
\begin{tabular}{l|ccccc}
\toprule
Data set & WSS & \ADIDA & iETS & NegBinGP & TweedieGP \\
\midrule
\rowcolor{backcolour} M5 & $0.10 \pm 0.03$ & $0.01 \pm 0.01$ & $0.76 \pm 0.07$ & $1.20 \pm 0.39$ & $2.12 \pm 0.49$ \\
OnlineRetail & $0.11 \pm 0.05$ & $0.01 \pm 0.01$ & $0.44 \pm 0.06$ & $0.32\pm 0.18$ & $0.35 \pm 0.22$ \\
\rowcolor{backcolour} Auto & $0.04 \pm 0.02$ & $0.01 \pm 0.01$ & $0.13 \pm 0.03$ & $0.20 \pm 0.06$ & $0.20 \pm 0.08$ \\
Carparts & $0.05 \pm 0.02$ & $0.01 \pm 0.01$ & $0.14 \pm 0.03$ & $0.17 \pm 0.11$ & $0.24 \pm 0.12$ \\ 
\rowcolor{backcolour} RAF & $0.05 \pm 0.03$ & $0.01 \pm 0.01$ & $0.20 \pm 0.03$ & $0.22 \pm 0.06$ & $0.30 \pm 0.07$ \\
\bottomrule
\end{tabular}
\caption{Mean and standard deviation of the time (seconds) for training the models and generating forecasts. Experiments are run on the CPU of a M3 MacBook Pro. }
\label{tab:comp_times}
\end{table*}

We report in Tab.~\ref{tab:comp_times}
the computational times \Rev{per time series} of each method, including training and generating the forecasts on time series of \Rev{each data set}. \Rev{Empirical quantiles are practically immediate and their execution times are not shown in the table.}
\ADIDA~and WSS are the fastest \Rev{methods}.

The computational time of our \Rev{GPs} is comparable with that of iETS. On the data sets containing the longest time series (M5), the average computational time 
of the GPs is about 2 seconds.
An exact GP implementation
would be rather slow on the time series of the M5 data set,
due to its cubic computational complexity 
in the number of observations $T$. Thanks to the variational approximation, the computational complexity of our models is instead quadratic in $T$ for $T \leq 200$. For longer time series the cost is capped by the variational formulation and becomes linear in $T$. See~\ref{sec:learning_variationalGP} for more details. 

The computational times of TweedieGP and NegBinGP are close; the fully parameterised Tweedie density only implies a small overhead compared to the negative binomial, \Rev{which is simple to evaluate}. 
On the M5 data set, NegBinGP is faster than TweedieGP as it often meets the early stopping condition. 

\subsection{Ablation study}
\label{subsec:likComparison}

We compare TweedieGP against a GP model trained with the approximate Tweedie likelihood with $A(y) = 1$ and $\phi = 1$ (implied by the Tweedie loss) and a TweedieGP trained with \Rev{no scaling}. 
We keep the same experimental setup of the previous section and we show the results in Tab.~\ref{tab:ablation_study}.

\begin{table}[!ht]
\centering
\begin{tabular}{ll|ccc}
\toprule
 \rotatebox[origin=c]{90}{Data set}& \rotatebox[origin=c]{90}{Metric}& \rotatebox[origin=c]{90}{\Rev{No scaling}} & \rotatebox[origin=c]{90}{\Rev{$A(y)=\phi=1$}} & \rotatebox[origin=c]{90}{\Rev{TweedieGP}} \\
\midrule
\greytabline & $\mathrm{sQ}_{0.5}$ & \textbf{1.75} & 1.76 & \textbf{1.75} \\
 & $\mathrm{sQ}_{0.8}$ & \textbf{1.45} & \textbf{1.45} & \textbf{1.45} \\
\greytabline & $\mathrm{sQ}_{0.9}$ & \textbf{1.26} & 1.29 & \textbf{1.26} \\
 & $\mathrm{sQ}_{0.95}$ & \textbf{1.16} & 1.23 & \textbf{1.16} \\
\greytabline & $\mathrm{sQ}_{0.99}$ & 1.17 & 1.38 & \textbf{1.16} \\
 & s\CRPS & \textbf{1.48} & 1.49 & \textbf{1.47} \\
 \greytabline \multirow[c]{-7}{*}{\rotatebox[origin=c]{90}{M5}}
& \RMSSE & 0.94 & 0.94 & \textbf{0.93}\\
\midrule

& $\mathrm{sQ}_{0.5}$ & 2.36 & 2.42 & \textbf{2.34} \\
\greytabline & $\mathrm{sQ}_{0.8}$ & 2.28 & 2.30 & \textbf{2.25} \\
 & $\mathrm{sQ}_{0.9}$ & 2.24 & 2.24 & \textbf{2.19} \\
\greytabline & $\mathrm{sQ}_{0.95}$ & 2.32 & 2.30 & \textbf{2.21} \\
 & $\mathrm{sQ}_{0.99}$ & 3.36 & 3.40 & \textbf{2.98} \\
\greytabline & s\CRPS & 2.33 & 2.37 & \textbf{2.29} \\
 \multirow[c]{-7}{*}{\rotatebox[origin=c]{90}{OnlineRetail}}
 & \RMSSE & \textbf{0.95} & 0.96 & \textbf{0.95} \\
\midrule

\greytabline & $\mathrm{sQ}_{0.5}$ & 1.19 & \textbf{1.17} & 1.19 \\ 
& $\mathrm{sQ}_{0.8}$ & 1.34 & 1.32 & \textbf{1.30} \\
\greytabline & $\mathrm{sQ}_{0.9}$ & 1.50 & 1.47 & \textbf{1.42} \\
 & $\mathrm{sQ}_{0.95}$ & 1.77 & 1.69 & \textbf{1.63} \\
\greytabline & $\mathrm{sQ}_{0.99}$ & 3.05 & \textbf{2.61} & \textbf{2.63} \\ 
 & s\CRPS & 1.30 & \textbf{1.27} & \textbf{1.27} \\
 \greytabline \multirow[c]{-7}{*}{\rotatebox[origin=c]{90}{Auto}}
 & \RMSSE & \textbf{0.81} & 0.82 & 0.82 \\
\midrule

 & $\mathrm{sQ}_{0.5}$ & \textbf{1.11} & \textbf{1.11} & \textbf{1.11} \\
\greytabline & $\mathrm{sQ}_{0.8}$ & \textbf{1.09} & 1.11 & \textbf{1.09} \\
& $\mathrm{sQ}_{0.9}$ & \textbf{1.13} & \textbf{1.13} & \textbf{1.13} \\
 \greytabline & $\mathrm{sQ}_{0.95}$ & \textbf{1.18} & \textbf{1.19} & 1.20 \\
& $\mathrm{sQ}_{0.99}$ & \textbf{1.56} & \textbf{1.59} & \textbf{1.55} \\
\greytabline & s\CRPS & \textbf{1.10} & \textbf{1.11} & \textbf{1.10} \\
\multirow[c]{-7}{*}{\rotatebox[origin=c]{90}{Carparts}} 
 & \RMSSE & \textbf{0.61} & \textbf{0.61} & \textbf{0.61} \\
 \midrule 
 
\greytabline & $\mathrm{sQ}_{0.5}$ & 1.01 & 1.03 & \textbf{1.00} \\
 & $\mathrm{sQ}_{0.8}$ & 1.08 & 1.20 & \textbf{1.02} \\
\greytabline & $\mathrm{sQ}_{0.9}$ & \textbf{1.15} & 1.21 & \textbf{1.15} \\
 & $\mathrm{sQ}_{0.95}$ & \textbf{1.24} & 1.27 & 1.26 \\
\greytabline & $\mathrm{sQ}_{0.99}$ & 2.36 & 2.28 & \textbf{2.09} \\
 & s\CRPS & 1.11 & 1.18 & \textbf{1.08} \\
\greytabline \multirow[c]{-7}{*}{\rotatebox[origin=c]{90}{RAF}}
 & \RMSSE & \textbf{0.60} & 0.61 & 0.61 \\
\bottomrule
\end{tabular}
\caption{Comparison of GP model with different likelihoods; all models use the RBF kernel.} \label{tab:ablation_study}
\end{table}

The accuracy of TweedieGP \Rev{worsens on unscaled data}. Due to numerical issues, the posterior mean of the Gaussian process does not grow over a certain \Rev{value}, which is problematic when demand \Rev{sizes} become high.

Finally, the performance of the approximated Tweedie likelihood with $A\Rev{(y)}=\phi=1$ confirms what discussed in Sec.~\ref{subsec:TweedieLossDensity}: the Tweedie loss might produce effective point forecasts by penalizing values far from zero, but the shorter tails negatively affect the estimate of the higher quantiles.

\section{Conclusions}
\label{sec:conclusion}
\Rev{The prediction of intermittent time series should be based on probability distributions rather than on point forecasts.
The most common probabilistic models are based on a latent variable
related to the demand size and to the probability of occurrence.
This paper proposes to model such a latent variable with a Gaussian Process, which quantifies the uncertainty on the latent variable and propagates it into the forecast distribution, thus obtaining more reliable prediction intervals.}

\Rev{We test two variants of GP with different likelihoods: the negative binomial and the Tweedie.
The two models perform similarly, but TweedieGP performs better on the high quantiles.
Our model is the first probabilistic model for intermittent time series which adopts the Tweedie likelihood, which is flexible but whose density is difficult to evaluate.
Hence we find it useful to publicly release our implementation of the Tweedie likelihood,
making it usable to train novel probabilistic models for intermittent time series in the future. 
}



\Rev{Our GP model uses only a}
RBF kernel; in future works, \Rev{we could} add also a periodic kernel
to model the effect of seasonality, \Rev{which is an open problem in intermittent time series \citep{FILDES20221283}. Adding a periodic kernel to the RBF already proved to be successful in smooth time series \citep{Corani_Benavoli_Zaffalon_2021}.}
We also leave for future works the comparison between local and global models for intermittent time series.



\section*{CRediT authorship contribution statement}

\textbf{Stefano Damato}: Writing – review \& editing, Writing original draft, Visualization, Validation, Software, Resources, Project administration, Methodology, Investigation, Formal analysis, Data curation, Conceptualization.
\textbf{Dario Azzimonti}:
Writing - review \& editing, Writing original draft, Visualization, Validation, Supervision, Resources, Project administration, Methodology, Investigation, Funding acquisition, Formal analysis, Data curation, Conceptualization.
\textbf{Giorgio Corani}: 
Writing - review \& editing, Writing original draft, Visualization, Validation, Supervision, Resources, Project administration, Methodology, Investigation, Funding acquisition, Formal analysis, Data curation, Conceptualization.

\section*{Declaration of competing interest}

The authors declare that they have no known competing financial interests or personal relationships that could have appeared to influence the work reported in this paper.

\section*{Acknowledgments}
This work is partially funded by the Swiss National Science Foundation (SNF), grant 200021\_212164/1.
This project has received funding from the European Union’s Horizon Europe Research and Innovation Framework under grant agreement No.101160720.

\bibliographystyle{elsarticle-harv}
\bibliography{refs}

\newpage
\appendix

\section{Learning a sparse variational GP}\label{sec:learning_variationalGP}

Recall that our forecasting model learns the posterior of the latent function $\mathbf{f}_{1:T}$ given the observations. The prior for a latent vector of length $T$ (training data length) is $p(\mathbf{f}_{1:T}) = \mathcal{N}(\mathbf{f}_{1:T} \mid \mathbf{0}, K_{T,T})$ and the joint distribution of data and latent variables is 
\begin{equation}
 p(\mathbf{y}_{1:T}, \mathbf{f}_{1:T}) = \prod_{i=1}^T \Rev{p_{\mathrm{lik}}}(y_i; \mathrm{softplus}(f_i), \Rev{\boldsymbol{\theta}_{\mathrm{lik}}}) \mathcal{N}(\mathbf{f}_{1:T} \mid \mathbf{0}, K_{T,T})
 \label{eq:jointTweedie}
\end{equation}
The posterior over latent function $p(\mathbf{f}_{1:T} \mid y_{1:T})$ is not available analytically therefore we need to approximate it. Moreover, in order to optimise the hyper-parameters of the model, we also need to approximate the marginal likelihood $p(\mathbf{y}_{1:T})$. We follow \citet{Hensman_Matthews_Ghahramani_2015} and use a sparse GP model with inducing points.

We proceed by augmenting our GP model with additional $m$ input-output pairs $\mathbf{Z}$, $\mathbf{u}$ that are distributed as the GP $f$, i.e. the joint distribution of the vector $(\mathbf{f}_{1:T},\mathbf{u})$ is $$p(\mathbf{f}_{1:T},\mathbf{u}) = \mathcal{N}\left(\begin{bmatrix}
 \mathbf{f}_{1:T} \\
 \mathbf{u}
\end{bmatrix} \ \Bigg| \ \mathbf{0}, \begin{bmatrix}
 K_{T,T} & K_{T,m} \\
 K_{m,T} & K_{m,m}
\end{bmatrix}\right),$$
where $K_{m,m}$ and $K_{T,m}$ are the covariance matrices resulting from evaluating the kernel $k$ at the input values. Note that the Gaussian assumption implies that $p(\mathbf{f}_{1:T} \mid \mathbf{u})$ is available analytically via Gaussian conditioning. 

The joint distribution of data and latent variables thus becomes $p(\mathbf{y}_{1:T}, \mathbf{f}_{1:T}, \mathbf{u}) \allowbreak = p(\mathbf{y}_{1:T} \mid \mathbf{f}_{1:T})p(\mathbf{f}_{1:T} \mid \mathbf{u})p(\mathbf{u})$. We consider the approximate distribution $q(\mathbf{u}) = \mathcal{N}(\mathbf{u} \mid \mathbf{m}, \mathbf{S})$, where $\mathbf{m}, \mathbf{S}$ are free parameters to be optimized. 

In variational inference we assume that the posterior is approximated by $p(\mathbf{f}_{1:T} \mid \mathbf{y}_{1:T}) \approx q(\mathbf{f}_{1:T} \mid \mathbf{y}_{1:T}) = \int p(\mathbf{f}_{1:T} \mid \mathbf{u})q(\mathbf{u}) d\mathbf{u}$. Since $q(\mathbf{u})$ is Gaussian we can solve the integral analytically. We can then bound the marginal log-likelihood $\log p(\mathbf{y}_{1:T})$ with the standard variational bound \citep{Hensman_Fusi_Lawrence_2013}
$$\log p(\mathbf{y}_{1:T}) \geq \mathbb{E}_{q(\mathbf{u})}[\log p(\mathbf{y}_{1:T} \mid \mathbf{u})] - \mathrm{KL}[q(\mathbf{u}) \mid p(\mathbf{u})],$$
which can be further bounded as 
\begin{equation}
\label{eq:ELBO}
 \log p(\mathbf{y}_{1:T}) \geq \mathbb{E}_{q(\mathbf{f})}[\log p(\mathbf{y}_{1:T} \mid \mathbf{f})] - \mathrm{KL}[q(\mathbf{u}) \mid p(\mathbf{u})]. 
\end{equation}

The right-hand side of eq.~\eqref{eq:ELBO} is the evidence lower bound (ELBO). This is a loss function that can be used to optimize the model hyper-parameters ($c, \Rev{\ell, \sigma^2}, \boldsymbol{\theta}_{\mathrm{lik}}$), the location of the inducing points ($\mathbf{Z}$) and the variational parameters ($\mathbf{m}, \mathbf{S}$). 

The KL part of the ELBO is available analytically, however the expectation part needs an implementation of the log-likelihood function. The expectation is then evaluated with Monte Carlo sampling by exploiting the fact that $q(\mathbf{f})$ is a Gaussian distribution with known parameters. 

We implemented the log-likelihood function $\log p(\mathbf{y}_{1:T} \mid \mathbf{f})$ for the Tweedie likelihood and then used GPyTorch \citep{Gardner_Pleiss_Weinberger_Bindel_Wilson_2018} for optimizing the ELBO. 

Given the optimized variational posterior approximation $q(\mathbf{f}_{1:T} \mid \mathbf{y}_{1:T})$, we can compute the predictive latent distribution $h$-step ahead as
\begin{equation}
 p(\mathbf{f}_{T+1:T+h} )= \int p(\mathbf{f}_{T+1:T+h} \mid \mathbf{f}_{1:T} ) q(\mathbf{f}_{1:T} \mid \mathbf{y}_{1:T})\mathrm{d}\mathbf{f}_{1:T},
 \label{eq:predLatent}
\end{equation}
note that this integral has an analytic solution because the distributions are both Gaussian therefore $\mathbf{f}_{T+1:T+h}$ is a multivariate Gaussian distribution with known mean and covariance; see, e.g., \citet{Hensman_Matthews_Ghahramani_2015} for detailed formulae. 

The predictive posterior for the observations is computed by \Rev{drawing} samples $\mathbf{\tilde{f}}^{(j)}$, $j=1, \ldots, N$, from the distribution in eq.~\eqref{eq:predLatent}. For each sample $\mathbf{\tilde{f}}^{(j)}$, we draw one sample from 
\begin{equation} 
p( \mathbf{y}_{T+1:T+h} \mid \mathbf{\tilde{f}}^{(j)}_{T+1:T+h}) = \prod_{i=T+1}^{T+h}\Rev{p_{\mathrm{lik}}} \left(y_i ; \mathrm{softplus}(\tilde{f}^{(j)}_i), \Rev{\boldsymbol{\theta}_{\mathrm{lik}}} \right) 
\label{eq:samplingpred}
\end{equation}
to obtain a sample from the predictive posterior.
\Rev{This procedure is applied for NegBinGP and TweedieGP by plugging in eq.~\eqref{eq:jointTweedie} and \eqref{eq:samplingpred} the appropriate likelihoods and parameters from Sec.~\ref{subsec:negBinLik} and Sec.~\ref{subsec:TweedieLik} respectively.}

In our experiments we choose the number of inducing points as follows. On short time series ($T \leq 200$), we use $T$ inducing points and initialize their locations in correspondence of the training inputs. 
On longer time series ($T > 200$) we use $200$ inducing points, sampling their initial locations from a multinomial distribution with $p(i)\propto\log \left( 1 + \frac{i}{T} \right)$, $i = 1, \ldots, T$.
Thus the recent observations, which are more relevant for forecasting, have higher probability of being chosen as initial location of the inducing points. Since the method has a quadratic cost in $m$, the number of inducing points, and it is linear in $T$, the size of the training set, the cost scales quadratically in $T$ for $T\leq 200$ and linearly in $T$ for $T>200$. . 


\section{Evaluation of the Tweedie density}

\subsection{Truncating the infinite summation}
\label{sec:truncating_sum}

In Sec.~\ref{subsec:tweedie_distr} we show that, given $\mu > 0$, $\phi > 0$ and $\rho \in (1,2)$ the Tweedie density is
\begin{equation*}
p(y \mid \mu, \phi, \rho) = \begin{cases} \exp(-\frac{\mu^{2-\rho}}{\phi(2-\rho)}) & \text{if} \ y=0; \\
A(y) \cdot \exp \left[ \frac1\phi \left( y \frac{\mu^{1-\rho}}{1-\rho} - \frac{\mu^{2-\rho}}{2-\rho} \right) \right] & \text{otherwise}
\end{cases}
\end{equation*}
where 
\begin{align}
A(y) &= \frac{1}{y}\sum_{j=1}^{\infty} \frac{y^{j\alpha}(\rho-1)^{-j\alpha }}{\phi^{j(1+\alpha)}(2-\rho)^jj!\Gamma(j\alpha)} \nonumber\\
&= \frac{1}{y}\sum_{j=1}^{\infty} V(j) \quad \text{with} \ \alpha = \frac{2-\rho}{\rho -1}
\label{eq:sum_A_alpha}
\end{align}
\cite{Dunn_Smyth_2005} evaluate $A(y)$ by approximating the infinite sum in eq.~\eqref{eq:sum_A_alpha} with a finite one, retaining its largest elements. We start by finding 
\[
\jmax = \argmax_{j \in \mathbb{N}} V(j),
\] 
which is the index of the largest component. To identify it, let 
\begin{equation}
z = \frac{y^{\alpha} (\rho-1)^{-\alpha}}{\phi^{1+\alpha}(2-\rho)}, 
\label{eq:z} 
\end{equation}
then $V(j) = \frac{z^j}{j!\Gamma(\alpha j)}$ and
\begin{equation} \log V(j) = j \log z - \log \Gamma \left( 1+j \right) - \log \Gamma \left( \alpha j \right). 
\label{eq:log_Vj}
\end{equation}
Stirling's approximation simplifies the evaluation of the Gamma function: $\Gamma(x+1) \simeq \sqrt{2\pi x} \cdot x^x \cdot e^{-x}$. Therefore 
\begin{equation}
\log \Gamma(x+1) \simeq \frac12 \log 2\pi + \frac12 \log x + x \log x -x .
\label{eq:stirling}
\end{equation}
Approximating $\Gamma(\alpha j)$ with $\Gamma(1+ \alpha j)$ in eq.~\eqref{eq:log_Vj} and using eq.~\eqref{eq:stirling} we have
\begin{align*}
\log V(j) \simeq &j\log z - \frac12 \log 2\pi -\frac12 \log j - j \log j + j - \frac12 \log 2\pi \\
&-\frac12 \log \alpha -\frac12 \log j - \alpha j \log \alpha - \alpha j \log j + \alpha j \\
= &j \left( \log z +( 1 + \alpha) - \alpha \log \alpha - (1+ \alpha )\log j \right) \\
&- \log 2\pi - \frac12 \log \alpha - \log j .
\end{align*}
Differentiating with respect to $j$ we obtain
\begin{align*}
 \frac{\partial \log V(j)}{\partial j} \simeq & \log z + (1+\alpha) - \alpha \log \alpha - (1+\alpha ) \log j - j(1+\alpha) \frac1j - \frac1j \\ 
 = & \log z - \log j - \alpha \log (\alpha j) - \frac1j
 \\ \simeq & \log z - \log j - \alpha \log (\alpha j) 
\end{align*}
which is a monotone decreasing function of $j$. For this reason, $\log V$ is a convex function (and therefore $V$ too). To identify its maximum point, we equate it to $0$ to get
\begin{equation}
(1+\alpha) \log \jmax = \log z - \alpha \log \alpha 
\label{eq:1_alpha_logjmax}
\end{equation}
and substituting $z$ from eq.~\eqref{eq:z} and $\alpha$ from eq.~\eqref{eq:sum_A_alpha}
\begin{align*}
 \jmax^{1+\alpha} 
 &= \frac{y^\alpha (\rho-1)^{-\alpha}}{ \phi^{1+\alpha}(2-\rho)} \cdot \frac{(2-\rho)^{-\alpha}}{(\rho-1)^{-\alpha}}
\end{align*}
and finally, since $\frac{\alpha}{1+\alpha} = 2-\rho$:
\begin{equation}
\jmax = \frac{y^{2-\rho}}{\phi (2-\rho)},
\label{eq:jmax}
\end{equation}
where it can be appropriately rounded to be a natural number. Substituting $\jmax$ into $V$ using eq.~\eqref{eq:1_alpha_logjmax}, it gives 
\begin{equation}
\log V(\jmax) = \jmax (1+\alpha) - \log 2 \pi - \log \jmax - \frac12 \log \alpha.
\label{eq:logVjmax}
\end{equation}
From here, we evaluate $V(j)$ for $j = \jmax+1, \jmax+2, \dots$ until a stopping index $j_U$ is found, such that
\[
 \frac{V(\jmax)}{V(j_U)} \geq e^{37}.
\]
To compute that, 
one can evaluate the difference between eqs.~\eqref{eq:log_Vj} and \eqref{eq:logVjmax} as
\begin{equation}
\begin{aligned}
 \log V(\jmax) - \log V(j_U) \simeq & (\jmax - j_U) C_W- \left( \jmax \left( 1 + \alpha\right) + 1 \right) \log \jmax \\ &+ \left( j_U \left( 1 + \alpha\right) + 1 \right) \log j_U
 \label{eq:logVmax_logV}
\end{aligned}
\end{equation}
where $C_W = \log z + (1+\alpha) -\alpha \log \alpha $ does not depend on $j$; such expression can be obtained efficiently. Similarly, one can identify $j_L$ such that 
\[
 \frac{V(\jmax)}{V(j_L)} \geq e^{37}
\]
going backwards to $\jmax-1, \jmax-2$, potentially stopping at $j=1$. 
The fact that $\log V$ is a convex function implies that the value of $V$ decreases exponentially on both sides.
Hence, the truncated part of the summation can be bounded with geometric sums:
\[
 \sum_{j=1}^{+\infty}V(j) - \sum_{j=j_L}^{j_U}V(j) \leq V\left(j_L - 1 \right) \frac{1-r_L^{j_L -1}}{1 - r_L} + V\left(j_U + 1 \right) \frac{1}{1 - r_U}
\]
where 
\[
r_L = \exp \left( \frac{\partial \log V(j)}{\partial j} \right) \biggr\rvert_{j = j_L -1} \quad \mathrm{and} \quad r_U = \exp \left( \frac{\partial \log V(j)}{\partial j} \right) \biggr\rvert_{j = j_U +1}.
\]
The \Rev{threshold} $e^{-37}$ has been proposed since $e^{-37} \simeq 8 \cdot 10^{-17}$ guarantees an appropriate precision using 64-bit floating points. See \citet{Dunn_Smyth_2005} for more. In practice, we are interested in computing the log-likelihood \[
\log p(y \mid \mu, \phi, \rho) = \log A(y) + \left[ \frac1\phi \left( y \frac{\mu^{1-\rho}}{1-\rho} - \frac{\mu^{2-\rho}}{2-\rho} \right) \right]
\]
which similarly requires to pass $A(y)$ trough the logarithm.
To implement the evaluation of $A(y)$, it sufficient to refer to eqs.~\eqref{eq:sum_A_alpha},~\eqref{eq:log_Vj},~\eqref{eq:jmax},~\eqref{eq:logVjmax}, and~\eqref{eq:logVmax_logV}.

\begin{table*}[h]
{\scriptsize
\begin{tabular}{ll|rrrrr|rrrrr|rrrrr|rrrrr}
\toprule
& $\phi$ & \multicolumn{5}{c}{0.5}& \multicolumn{5}{c}{1.0} & \multicolumn{5}{c}{2.0} & \multicolumn{5}{c}{5.0} \\
 & $\rho$ & \rotatebox{90}{1.01} & \rotatebox{90}{1.1} & \rotatebox{90}{1.2} & \rotatebox{90}{1.3} & \rotatebox{90}{1.5} & \rotatebox{90}{1.01} & \rotatebox{90}{1.1} & \rotatebox{90}{1.2} & \rotatebox{90}{1.3} & \rotatebox{90}{1.5} & \rotatebox{90}{1.01} & \rotatebox{90}{1.1} & \rotatebox{90}{1.2} & \rotatebox{90}{1.3} & \rotatebox{90}{1.5} & \rotatebox{90}{1.01} & \rotatebox{90}{1.1} & \rotatebox{90}{1.2} & \rotatebox{90}{1.3} & \rotatebox{90}{1.5} \\
\midrule
 \greytabline & 0.1 & 2 & 3 & 5 & 7 & 14 & 2 & 2 & 4 & 6 & 10 & 2 & 2 & 3 & 5 & 8 & 2 & 2 & 3 & 4 & 6 \\
& 0.2 & 2 & 4 & 6 & 9 & 15 & 2 & 3 & 5 & 7 & 11 & 2 & 2 & 4 & 5 & 9 & 2 & 2 & 3 & 4 & 7 \\
 \greytabline & 0.5 & 3 & 6 & 9 & 12 & 18 & 2 & 4 & 6 & 8 & 14 & 2 & 3 & 5 & 6 & 10 & 2 & 2 & 4 & 5 & 8 \\
& 0.8 & 4 & 7 & 11 & 14 & 21 & 2 & 5 & 8 & 10 & 15 & 2 & 3 & 5 & 7 & 11 & 2 & 2 & 4 & 5 & 8 \\
\greytabline y & 1 & 5 & 9 & 12 & 14 & 21 & 3 & 6 & 9 & 11 & 16 & 2 & 4 & 6 & 8 & 12 & 2 & 3 & 4 & 6 & 9 \\
& 1.5 & 5 & 10 & 14 & 17 & 24 & 4 & 7 & 9 & 12 & 18 & 2 & 5 & 7 & 9 & 14 & 2 & 3 & 5 & 6 & 9 \\
\greytabline & 2 & 5 & 12 & 16 & 19 & 26 & 5 & 8 & 11 & 14 & 18 & 3 & 6 & 8 & 10 & 14 & 2 & 3 & 5 & 7 & 10 \\
& 5 & 7 & 19 & 24 & 27 & 33 & 6 & 12 & 16 & 18 & 23 & 5 & 9 & 11 & 13 & 17 & 2 & 5 & 7 & 8 & 11 \\
\greytabline \multirow{9}{*}{\rotatebox{90}{$y$}} & 10 & 9 & 25 & 32 & 36 & 40 & 7 & 18 & 21 & 24 & 28 & 6 & 12 & 14 & 16 & 20 & 4 & 7 & 9 & 11 & 14 \\
\bottomrule
\end{tabular}
}
\caption{$j_U - j_L + 1$ for different choices of $y$, $\phi$ and $\rho$. This is the amount of terms used in the approximation of the summation in $A(y)$. It is an increasing function with respect to $y$ and $\rho$, decreasing with respect to $\phi$.}
\label{tab:summation_terms}
\end{table*}

\begin{figure}[ht]
 \centering
 \includegraphics[width=1\linewidth]{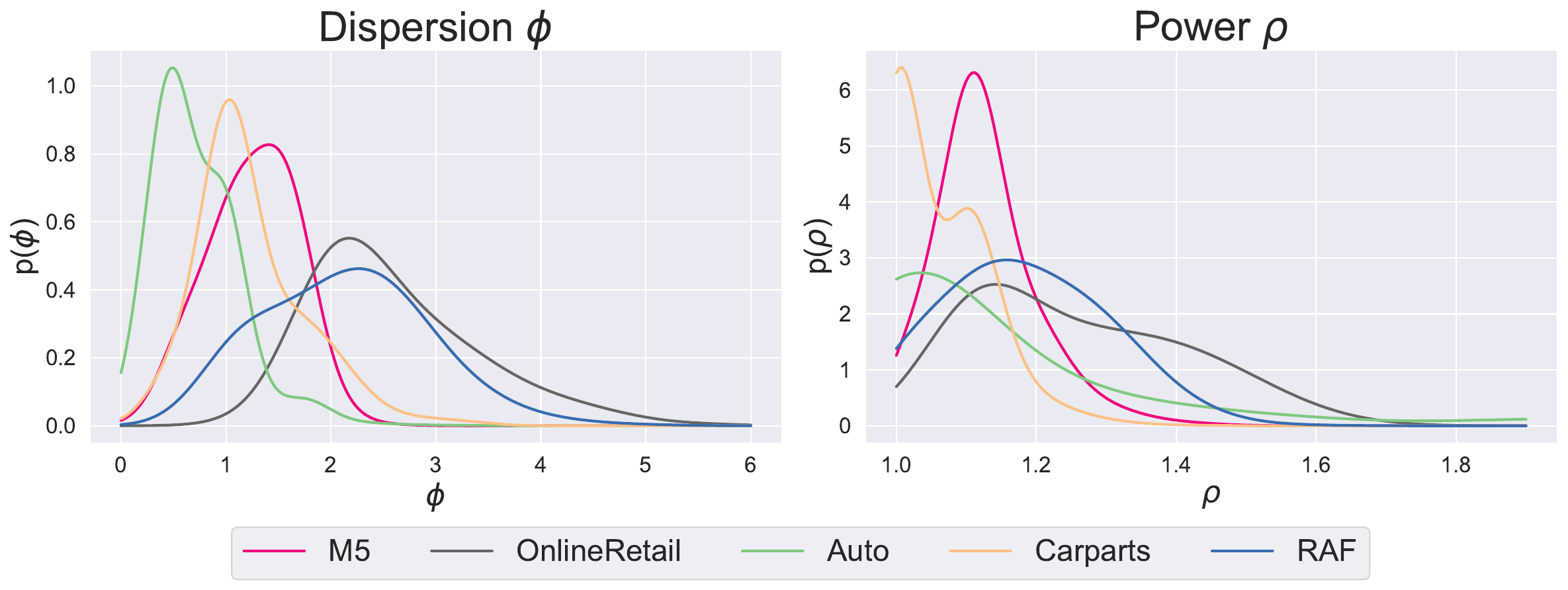}
 \caption{The values of $\phi$ and $\rho$ on the fitted TweedieGP on different data sets. Density curves have been computed via kernel density estimation.}
 \label{fig:params_densities}
\end{figure}

Table~\ref{tab:summation_terms} shows the amount of terms in the approximation of the summation $A(y)$ is generally affordable; indeed scaling the data by median demand \Rev{size}, the value of $y$ typically lies between 0.5 and 2, while the density plots from Fig.~\ref{fig:params_densities} show that often $\phi \in (0.5, 5)$ with mode around 1.5, and $\rho \in (1, 1.5)$, with mode around $1.1$.

\subsection{Comparison with Tweedie loss}\label{sec:comparison_tweedie_loss}

In order to node the crucial role of $A(y)$ in fitting appropriately a Tweedie likelihood to the data, consider again the formula of the Tweedie loss, characterised by $A(y) = \phi = 1$: in this simplified version, the implied likelihood in eq.~\eqref{eq:tw_loss_p}
is no longer a probability distribution, as it does not integrate up to $1$. Therefore we determine $c := c(\mu, \phi, \rho)$ such that \begin{equation*}
 \int_0^{+\infty} c \cdot \exp \left( y \frac{\mu^{1-\rho}}{1-\rho} - \frac{\mu^{2-\rho}}{2-\rho}\right) dy = 1,
\end{equation*} and we include $c = \frac{\mu^{1-\rho}}{\rho -1}\exp \left( \frac{\mu^{2-\rho}}{2-\rho} \right)$ to obtain
\begin{equation*}
\tilde{p}(y \mid \mu, \rho) = \frac{\mu^{1-\rho}}{\rho -1} \exp \left( -y \frac{\mu^{1-\rho}}{\rho-1} \right),
\end{equation*}
which is the density function of a negative exponential distribution with parameter $ \frac{\mu^{1-\rho}}{\rho-1}$. In this perspective, the remainder term of the loss can be interpreted as a joint prior distribution on $\mu$ and $\rho$ which prevents the mean from growing too large. 

\begin{figure}[!h]
 \centering
\includegraphics[width=1\linewidth]{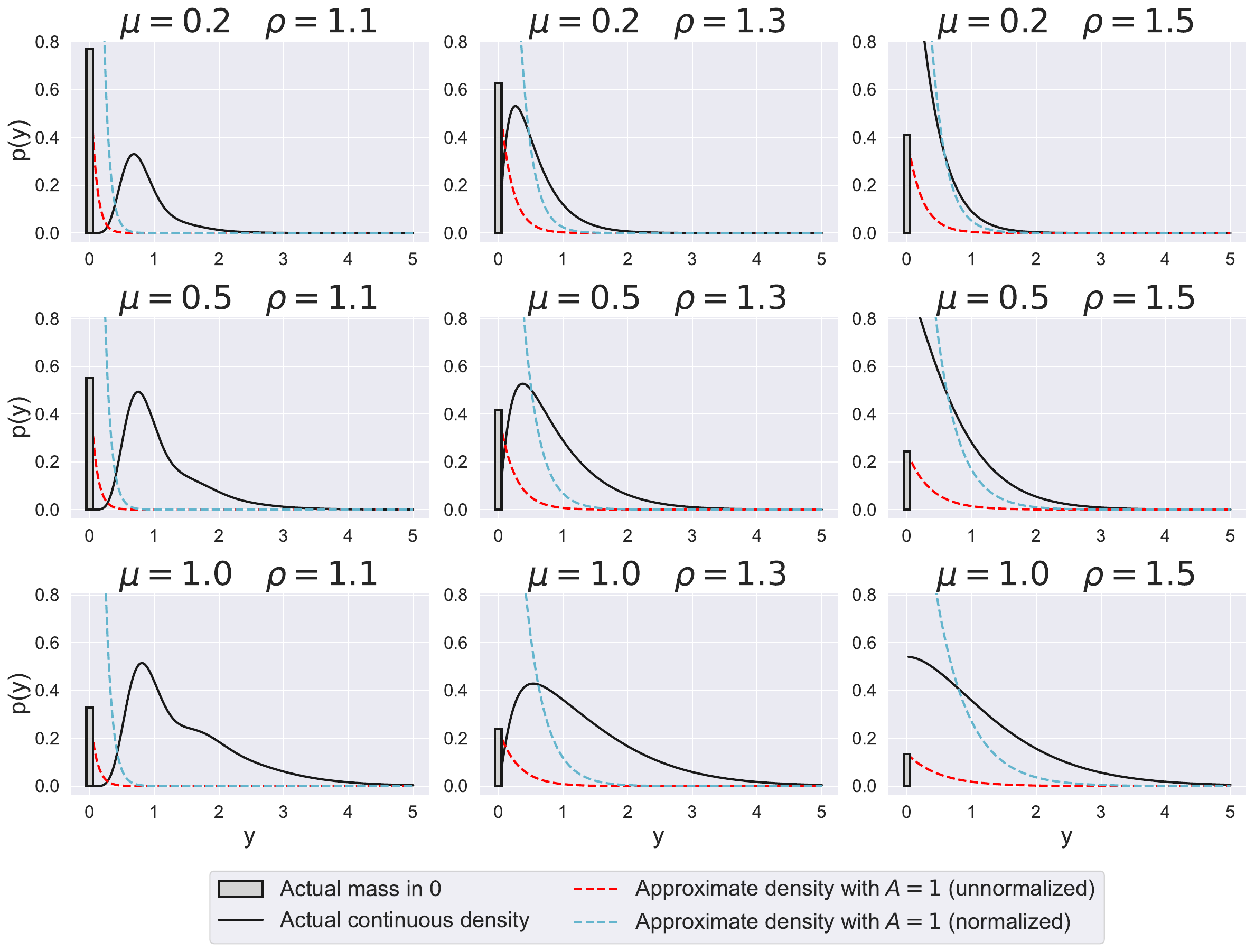}
 \caption{In this plot, $\phi=1$. In black, the Tweedie likelihood; the unnormalized density and its normalized version are the dashed lines in red and cyan respectively.}
\label{fig:comparison_tweedie_expanded}
\end{figure}

However, the distribution implied by the Tweedie loss is no longer bimodal, as shown in Fig.~\ref{fig:comparison_tweedie_expanded}, and empirical results show that having a \Rev{mass spike in zero} comes with the constraint of having short tails. For this reason its performance is not satisfactory on high quantiles, despite the use of the median demand scaling.

\section{Alternative results}
\label{sec:res_1.32}
\begin{table}[!ht]
 \centering
 \begin{tabular}{ll|cccccc}
 \toprule
 \rotatebox[origin=c]{90}{Data set} & \rotatebox[origin=c]{90}{Metric} & \rotatebox[origin=c]{90}{EmpQuant} & \rotatebox[origin=c]{90}{WSS} & \rotatebox[origin=c]{90}{\ADIDA} & \rotatebox[origin=c]{90}{iETS} & \rotatebox[origin=c]{90}{NegBinGP} & \rotatebox[origin=c]{90}{TweedieGP} \\ 
 \midrule 
 \greytabline & $\mathrm{sQ}_{0.5}$ & 1.88 & 1.87 & 1.88 & 1.83 & \textbf{1.78} & 1.79 \\
 & $\mathrm{sQ}_{0.8}$ & 1.74 & 1.77 & \textbf{1.46} & 1.65 & 1.48 & 1.47 \\
\greytabline & $\mathrm{sQ}_{0.9}$ & 1.57 & 1.65 & 1.35 & 1.56 & 1.30 & \textbf{1.27} \\
& $\mathrm{sQ}_{0.95}$ & 1.39 & 1.51 & 1.37 & 1.60 & 1.19 & \textbf{1.17} \\
\greytabline & $\mathrm{sQ}_{0.99}$ & 1.25 & 1.40 & 1.69 & 2.25 & \textbf{1.17} & \textbf{1.17} \\
 & s\CRPS & 1.73 & 1.75 & 1.57 & 1.71 & 1.51 & \textbf{1.50} \\
\greytabline \multirow[c]{-7}{*}{\rotatebox[origin=c]{90}{M5}}
 & \RMSSE & 0.98 & 1.00 & \textbf{0.94} & \textbf{0.94} & 0.95 & \textbf{0.94} \\
\midrule 

 & $\mathrm{sQ}_{0.5}$ & 2.35 & 2.35 & 2.55 & 2.36 & \textbf{2.34} & 2.35 \\
\greytabline & $\mathrm{sQ}_{0.8}$ & 2.34 & 2.33 & 2.29 & 2.40 & \textbf{2.25} &\textbf{2.26} \\
 & $\mathrm{sQ}_{0.9}$ & 2.31 & 2.29 & 2.23 & 2.51 & \textbf{2.19} & \textbf{2.19} \\ 
\greytabline & $\mathrm{sQ}_{0.95}$ & 2.33 & 2.30 & 2.35 & 2.75 & 2.24 & \textbf{2.22} \\
 & $\mathrm{sQ}_{0.99}$ & 3.02 & \textbf{2.89} & 4.09 & 4.36 & 3.10 & 2.99 \\
\greytabline & s\CRPS & 2.36 & 2.35 & 2.44 & 2.51 & \textbf{2.30} & \textbf{2.30} \\
 \multirow[c]{-7}{*}{\rotatebox[origin=c]{90}{OnlineRetail}} 
& \RMSSE & 0.97 & 0.98 & \textbf{0.94} & \textbf{0.94} & \textbf{0.94} & 0.95 \\
\midrule 

\greytabline & $\mathrm{sQ}_{0.5}$ & 1.15 & 1.36 & 1.19 & 1.17 & \textbf{1.13} & \textbf{1.13} \\
 & $\mathrm{sQ}_{0.8}$ & 1.30 & 1.53 & 1.34 & 1.39 & \textbf{1.28} & \textbf{1.28} \\
\greytabline & $\mathrm{sQ}_{0.9}$ & 1.48 & 1.65 & 1.59 & 1.54 & \textbf{1.42} & \textbf{1.42} \\
 & $\mathrm{sQ}_{0.95}$ & 1.84 & 1.84 & 2.21 & 1.76 & \textbf{1.64} & \textbf{1.62} \\
\greytabline & $\mathrm{sQ}_{0.99}$ & 4.32 & 2.57 & 7.51 & 3.28 & \textbf{2.41} & 2.56 \\
 & s\CRPS & 1.26 & 1.47 & 1.37 & 1.31 & \textbf{1.23} & \textbf{1.23} \\
\greytabline \multirow[c]{-7}{*}{\rotatebox[origin=c]{90}{Auto}}
& \RMSSE & \textbf{0.78} & 0.91 & 0.79 & \textbf{0.78} & \textbf{0.78} & \textbf{0.78} \\
\midrule 

 & $\mathrm{sQ}_{0.5}$ & 1.13 & 1.15 & 1.20 & \textbf{1.10} & \textbf{1.10} & 1.11 \\
\greytabline & $\mathrm{sQ}_{0.8}$ & 1.18 & 1.33 & 1.14 & 1.15 & \textbf{1.09} & \textbf{1.09} \\
 & $\mathrm{sQ}_{0.9}$ & 1.25 & 1.45 & 1.16 & 1.24 & 1.16 & \textbf{1.13} \\
\greytabline & $\mathrm{sQ}_{0.95}$ & 1.32 & 1.63 & 1.34 & 1.46 & 1.23 & \textbf{1.20} \\
 & $\mathrm{sQ}_{0.99}$ & 1.86 & 1.99 & 3.72 & 3.04 & 1.64 & \textbf{1.55} \\
\greytabline & s\CRPS & 1.19 & 1.29 & 1.25 & 1.18 & \textbf{1.11} & \textbf{1.10} \\
 \multirow[c]{-7}{*}{\rotatebox[origin=c]{90}{Carparts}} 
& \RMSSE & 0.66 & 0.78 & \textbf{0.60} & \textbf{0.59} & 0.61 & 0.61 \\
 \midrule 
 
\greytabline & $\mathrm{sQ}_{0.5}$ & \textbf{1.00} & \textbf{1.00} & 1.37 & \textbf{1.00} & \textbf{1.00} & \textbf{1.00} \\
 & $\mathrm{sQ}_{0.8}$ & \textbf{1.00} & 1.01 & 1.22 & \textbf{1.00} & 1.01 & 1.02 \\
\greytabline & $\mathrm{sQ}_{0.9}$ & 1.10 & 1.16 & 1.17 & \textbf{1.07} & 1.09 & 1.15 \\
 & $\mathrm{sQ}_{0.95}$ & \textbf{1.24} & 1.43 & 1.35 & 1.38 & \textbf{1.24} & 1.26 \\
\greytabline & $\mathrm{sQ}_{0.99}$ & 2.12 & 2.21 & 3.79 & 3.89 & 2.16 & \textbf{2.09} \\
 & s\CRPS & \textbf{1.06} & 1.09 & 1.38 & 1.12 & \textbf{1.06} & 1.08 \\
\greytabline \multirow[c]{-7}{*}{\rotatebox[origin=c]{90}{RAF}} 
& \RMSSE & 0.61 & 0.65 & 0.61 & \textbf{0.59} & 0.60 & 0.60 \\
 \bottomrule
\end{tabular}
 \caption{Same results as in Tab.~\ref{tab:s_bs}, but only considering time series with ADI $> 1.32$.}
 \label{tab:res_1.32}
\end{table}

\Revnew{We report in Tab.~\ref{tab:res_1.32} the aggregated metrics considering only time series with ADI $> 1.32$. On RAF data set, the scores remain completely unchanged. On M5 and Carparts, the results of all models undergo minimal changes that do not affect the conclusions drawn above. 
On OnlineRetail, the only noticeable difference is in the performance of WSS on the highest quantile, where it matches that of TweedieGP.}

\Revnew{On the Auto data set, whose series with ADI $> 1.32$ are 1227, GPs show a slight overall improvement. The performance of WSS and iETS, which model separately demand size and occurrence, generally worsens, especially on the higher quantiles. The results of the empirical quantiles and \ADIDA~improve on the lower quantiles, but worsen on the higher ones.}

\section{Data and code availability}\label{sec:reproducibility}

The data sets used in the experiments are downloadable by running the files located in the ``data'' folder at the GitHub page of the project, \url{https://github.com/StefanoDamato/TweedieGP}. The code for reproducing the experiments, tables, and plots in this paper is fully available too. In particular, our models are implemented in the
 \texttt{intermittentGP} class. It has methods to build, train and make predictions under the specification of the likelihood, the scaling and different training hyper-parameters, such as the number of epochs or the learning rate of the optimizer. The ``tutorial'' folder contains a simple usage example for our models. 

The implementation of our GP models is based on GPyTorch~\citep{Gardner_Pleiss_Weinberger_Bindel_Wilson_2018}. Our
contribution to public libraries is given by an implementation of the \texttt{Tweedie} class in the
\texttt{distributions} module of PyTorch~\citep{pytorch_2019}, and the classes \texttt{TweedieLikelihood} and
\texttt{NegativeBinomialLikelihood} in the \texttt{likelihoods} module of GPyTorch. We will contribute to those
packages with pull requests. 

\end{document}